\documentclass[letterpaper]{article} 
\usepackage{aaai25}
\usepackage{times}  
\usepackage{helvet}  
\usepackage{courier}  
\usepackage[hyphens]{url}  
\usepackage{graphicx} 
\usepackage{natbib}  
\usepackage{caption} 
\usepackage{algorithmic}
\usepackage[ruled,lined,linesnumbered]{algorithm2e}
\usepackage{multirow}
\usepackage{booktabs} 
\usepackage{amsmath}
\usepackage{amssymb}
\usepackage{amsxtra}
\usepackage{mathtools}
\usepackage{cleveref}
\usepackage{subcaption}
\usepackage{makecell}
\usepackage{amsfonts}       
\usepackage{nicefrac}       
\renewcommand{\vec}[1]{\boldsymbol{#1}}
\usepackage{xcolor}
\usepackage{newfloat}
\usepackage{listings}
\DeclareCaptionStyle{ruled}{labelfont=normalfont,labelsep=colon,strut=off} 
\usepackage{color}  
\nocopyright

\title{MergeOcc: Bridge the Domain Gap between Different LiDARs for Robust Occupancy Prediction}
\author{
    Zikun Xu\textsuperscript{\rm 1},
    Jianqiang Wang\textsuperscript{\rm $1_*$},
    Shaobing Xu\textsuperscript{\rm $1$}\thanks{Corresponding authors: shaobxu@tsinghua.edu.cn, wjqlws@tsinghua.edu.cn.} 
}
\affiliations{
    \textsuperscript{\rm 1}School of Vehicle and Mobility\\

    Tsinghua University, Beijing 100084, China\\
}

\usepackage{bibentry}

\begin{document}

\maketitle

\begin{abstract}
LiDAR-based 3D occupancy prediction algorithms evolved rapidly alongside the emergence of large datasets. 
Nevertheless, the potential of existing diverse datasets remains underutilized as they kick in individually. Models trained on a specific dataset often suffer considerable performance degradation when deployed to real-world scenarios or datasets involving disparate LiDARs.
This paper aims to develop a generalized model called MergeOcc, to simultaneously handle different LiDARs by leveraging multiple datasets.
The gaps among LiDAR datasets primarily manifest in geometric disparities and semantic inconsistencies. Thus, MergeOcc incorporates a novel model featuring a geometric realignment module and a semantic label mapping module to facilitate multiple datasets training (MDT). 
The effectiveness of MergeOcc is validated through experiments on two prominent datasets for autonomous vehicles: OpenOccupancy-nuScenes and SemanticKITTI. The results demonstrate its enhanced robustness and remarkable performance across both types of LiDARs, outperforming several SOTA multi-modality methods. 
Notably, despite using an identical model architecture and hyper-parameter set, MergeOcc can significantly surpass the baseline due to its exposure to more diverse data.
MergeOcc is considered the first cross-dataset 3D occupancy prediction pipeline that effectively bridges the domain gap for seamless deployment across heterogeneous platforms.
\end{abstract}

%

\section{Introduction}
\label{sec:intro}
\begin{figure}[t]
\centering
\includegraphics[width=6.8cm,height=5.5cm]{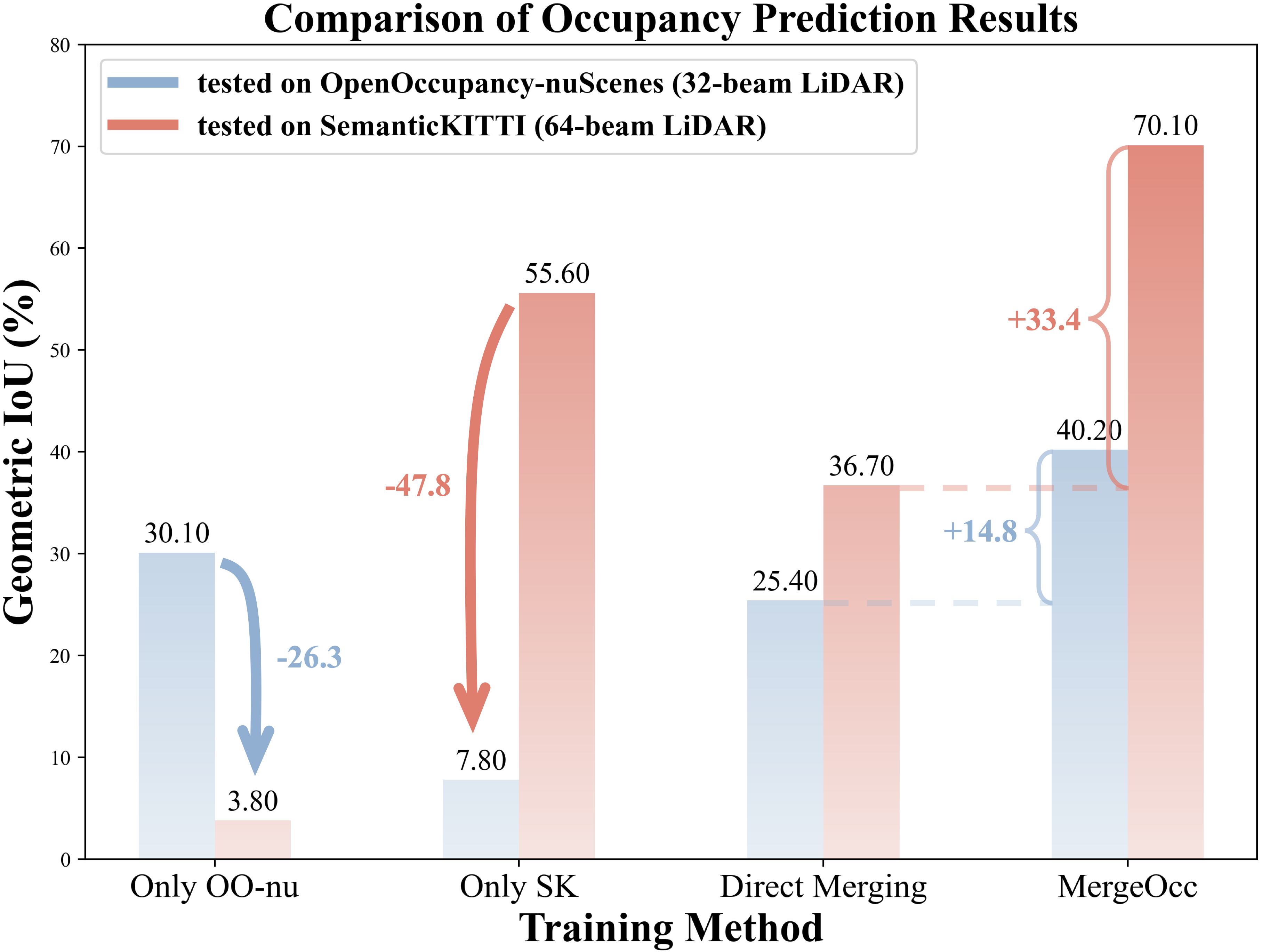}
\caption{Challenges of cross-LiDAR issue: 1) Only OO-nu and Only SK refer to the baseline model trained on each individual dataset. 2) Direct Merging represents training the same model on the simply merged dataset. 3) MergeOcc denotes the proposed method trained on these two datasets.}
\label{fig:drop}
\vspace{-0.4cm}
\end{figure}

With the capability to acquire precise geometric information of entire scenes, LiDAR becomes an indispensable sensor for many autonomous vehicles.
Current LiDAR-based occupancy prediction models mostly adhere to the paradigm of training and testing within a single dataset, thus constraining the source data to a limited domain. However, deploying dataset-specific models directly onto platforms or datasets equipped with different LiDARs often incurs notable performance degradation, as shown in \cref{fig:drop}, rendering that this paradigm fails to yield a robust and generalized perception model. 
This cross-LiDAR issue arises from substantial domain shifts between LiDARs. In terms of 3D data, the gaps manifest in two principal dimensions: \textbf{1)} significant divergence in point cloud density stemming from variations in LiDAR beams and angular resolutions, as detailed in ~\cref{tab1}, and \textbf{2)}, diversified scenarios and inconsistent semantic label spaces caused by disparate taxonomies.

Traditionally, addressing these challenges involves annotating data from the target domain followed by model fine-tuning. However, annotating a frame of LiDAR point cloud consumes approximately $10\times$ more time than annotating an image~\cite{xiaoSurveyLabelEfficientDeep2023}. Fully leveraging all available 3D LiDAR data could save the expenditure of time and resources. Furthermore, fine-tuning the model with a new dataset often results in \textbf{``catastrophic forgetting"}~\cite{kirkpatrickOvercomingCatastrophicForgetting2017}, where the model loses previously acquired knowledge. A question naturally arises: can these limitations be alleviated by bridging the gap between different LiDARs to obtain a more robust and generalized model?

\begin{table*}[t]
\small
\centering
\setlength{\tabcolsep}{0.4  mm}
{
\begin{tabular}{c|ccc|c|c|c|c}
\hline
Datasets & Beam & HAR & VFOV & Point Range & GT Range & GT shape & Collection Location \\
\hline
\multirow{3}{*}{SemanticKITTI
} & \multirow{3}{*}{64} & \multirow{3}{*}{$0.08^\circ$} & \multirow{3}{*}{[$-23.6^\circ$, $3.2^\circ$]} & L=[-72.0, 72.0]m 
& L=[0, 51.2]m
& \multirow{3}{*}{[256, 256, 32]} & \multirow{3}{*}{Germany}\\
&  &  &  & W=[-72.0, 72.0]m  &  W=[-25.6, 25.6]m & \\
&  &  &  & H=[-3.4, 3.0]m   & H=[-3.4, 3.0]m  &\\
\hline
\multirow{3}{*}{OpenOccupancy-nuScenes
} & \multirow{3}{*}{32} & \multirow{3}{*}{$0.08^\circ-0.33^\circ$} & \multirow{3}{*}{[$-30.0^\circ$, $10.0^\circ$]} & L=[-51.2, 51.2]m  
& L=[-51.2, 51.2]m
& \multirow{3}{*}{[512, 512, 40]} & \multirow{3}{*}{USA/Singapore} \\
&  &  &  & W=[-51.2, 51.2]m  & W=[-51.2, 51.2]m &\\
&  &  &  & H=[-5.0, 3.0]m  & H=[-5.0, 3.0]m & \\
\bottomrule
\end{tabular}
}
\caption{Differences of typical 3D LiDAR datasets for autonomous driving. HAR denotes horizontal angular resolution, VFOV denotes vertical field of view, and L, W, and H represent the length, width, and height of LiDAR range, respectively. }
\label{tab1}
\vspace{-0.2cm}
\end{table*}

Motivated by the aforementioned challenges, this paper introduces MergeOcc, a pioneering 3D occupancy prediction framework designed to simultaneously handle multiple types of LiDARs. Unlike prevailing LiDAR-based perception algorithms~\cite{wangOpenOccupancyLargeScale2023,zuoPointOccCylindricalTriperspective2023,weiSurroundOccMultiCamera3D2023}that primarily focus on developing a novel framework trained and validated on a singular dataset, we endeavor to furnish a methodology that enables existing 3D occupancy prediction models to leverage broader and more diverse LiDAR data, thereby enhancing model performance when directly deployed to real-world scenarios and reducing time and cost during migration across LiDARs. 

The contributions of this paper are as follows:
\begin{itemize}
\item To the best of our knowledge, MergeOcc stands as the first complete approach enabling one single model to simultaneously handle different LiDARs for 3D dense-perception tasks. MergeOcc features a geometric realignment module to capture unique statistical characteristics of each LiDAR within the model, alongside a semantic label mapping module to harmonize the label space across disparate datasets. These components bridge the gaps between disparate LiDARs, allowing the model to learn from broader data, thus improving its adaptability and performance across heterogeneous data domains.
\item We conduct extensive experiments on two prominent occupancy prediction datasets, OpenOccupancy-nuScenes and SemanticKITTI, to evaluate the efficacy of the proposed method in addressing domain gaps. MergeOcc exhibits exceptional robustness and performance with both types of LiDARs, achieving up to $33.4\%$ {IoU} and $17.2\%$ mIoU improvements compared to the D.M. method. Additionally, MergeOcc surpasses SOTA LiDAR and multi-modality methods by a commendable margin. 
\item Despite using an identical hyper-parameter set and architecture, MergeOcc can significantly outperform the baseline due to the superiority of the MDT paradigm, warning of the insufficient capacity of current LiDAR datasets.
\end{itemize}

\section{Related Works}
\label{sec:rw}

\subsection{3D Occupancy Prediction}
3D semantic occupancy prediction involves transforming real-world scenarios into fine-grained dense semantic occupancy grids, which offer a comprehensive depiction of the environment. This task is crucial for autonomous vehicles yet still requires further in-depth exploration.

Recently, Vision-centric methods have gained prominence. Wang et al.~\cite{wangOpenOccupancyLargeScale2023} introduced the pioneering benchmark OpenOccupancy-nuScenes for surrounding semantic occupancy perception in driving scenarios. SurroundOcc~\cite{weiSurroundOccMultiCamera3D2023} proposed a multi-camera-based 3D occupancy prediction pipeline, with training supervised by LiDAR-generated occupancy ground truths. TPVFormer~\cite{huangTriPerspectiveViewVisionBased2023} developed a novel tri-perspective view representation (TPV), which deviates from traditional voxel-based approaches, utilizing three orthogonal 2D planes to effectively model the 3D scene. 
 
In contrast, LiDAR point clouds contain inherently richer and more precise geometric information than images, and LiDAR-based methods deserve more exploration. 
Several early studies~\cite{xiaSCPNetSemanticScene2023,chengS3CNetSparseSemantic2021, yanSparseSingleSweep2021} conducted on SemanticKITTI explored the potential of LiDAR-based semantic scene completion and demonstrated commendable performance. PointOcc~\cite{zuoPointOccCylindricalTriperspective2023} transforms point clouds into TPV space and leverages a 2D image backbone to model detailed 3D structures, achieving comparable performance to voxel-based methods on the OpenOccupancy-nuScenes dataset. 

Nevertheless, all the aforementioned works adhere to the single-dataset training and testing paradigm. Consequently, they are susceptible to significant performance degradation in deployment across disparate LiDARs.

\subsection{Multiple Datasets Training (MDT)}
Training across multiple datasets is a proven strategy to enhance model robustness. 
Given the similar grid structure of images, the primary challenge in vision fields lies in unifying label space to ensure the consistency of semantic outputs. Early work attempted manually crafting the taxonomy~\cite{zhaoObjectDetectionUnified2020, lambertMSegCompositeDataset2020}, which are both labor-intensive and error-prone. 
With the development of large language models(LLM), differing from previous works, BigDetection ~\cite{caiBigDetectionLargescaleBenchmark2022} harnessed the power of BERT~\cite{devlinBERTPretrainingDeep2019} to construct an initial unified label space across datasets, followed by manual verification to finalize the taxonomy. However, these methods still require extensive human annotation, rendering them non-scalable to a larger scope. To address this limitation, Zhou et al.~\cite{zhouSimpleMultidatasetDetection2022} introduced a fully automatic approach based on combinatorial optimization to unify the output space of a multi-dataset detector by leveraging intrinsic image properties and similarity to derive the final taxonomy, offering a more scalable approach.

Unlike the consistency observed in the vision domain, 3D LiDAR data exhibits substantial disparities at both the geometric and semantic levels. Research on training with multiple LiDAR datasets remains limited. Uni3D ~\cite{zhangUni3DUnifiedBaseline2023} stood out as the pioneering effort focused on 3D multi-dataset object detection but focused exclusively on three key categories within autonomous driving scenarios.

In contrast to Uni3D, this paper addresses the dense prediction task on 3D occupancy grids using distinct training strategies and incorporates all available labels to construct a more comprehensive semantic space.

\begin{figure*}[!hbt]
\centering
\includegraphics[width=0.9\textwidth]{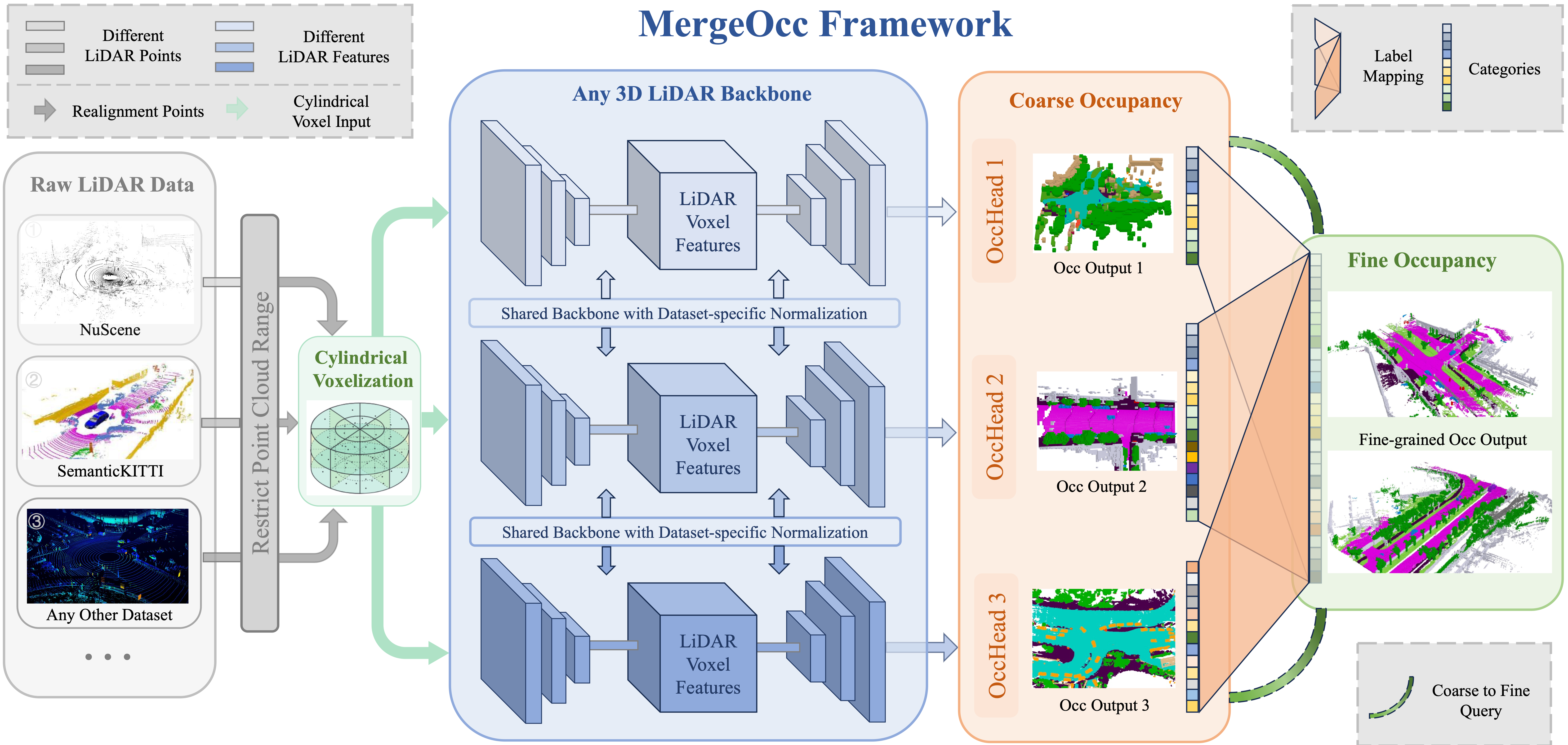}
\caption{The framework of MergeOcc including: 1) point range alignment, 2) cylindrical voxelization,  3) shared 3D backbone with dataset-specific normalization, 4) dataset-specific occupancy heads, 5) semantic label mapping, and 6) coarse to fine stage.}
\label{fig:framework}
\end{figure*}

\section{Method}
\label{sec:Method}
\subsection{Preliminary}
\label{sec:Preliminary}
\subsubsection{Problem Setting}
In the context of semantic occupancy prediction from a single dataset, the task involves processing an input frame of LiDAR points $\hat{X}_i \in \mathbb{R}^{N \times 3}$ to predict the corresponding occupancy labels $\mathcal{F}(\hat{X}_i) \in \mathbb{R}^{H \times W \times Z}$, where $H$, $W$, and $Z$ represent the volumetric dimensions of the entire scene. 
Training an occupancy prediction model on a single dataset usually follows a simple recipe:
minimize a total loss $\ell$, which often includes elements such as cross-entropy loss, semantic mIoU, and geometric IoU, across a set of  sampled point clouds $\hat X$ and its corresponding annotated occupancy ground truth $\hat L$ from the dataset $\mathcal{D}_i$:
\begin{equation}
    \min_{\Theta} \mathbb{E}_{(\hat X, \hat L) \in D_i} \left[\ell(\mathcal{F}(\hat X; \Theta), \hat L)\right]
\end{equation}
\subsubsection{Multiple Datasets Training}
Consider a dataset characterized by a joint probability distribution $P_{XL}$ over the input point cloud and label space $\mathcal{X} \times \mathcal{L}$. In the MDT paradigm, we denote the number of datasets ${\mathcal{D}_i={(\mathbf{x}^{(i)}, \mathbf{l}^{(i)})}}_{i=1}^{N}$ by $N$, where each $\mathcal{D}_i$ is associated with a unique data distribution $P_{XL}^{i}$. The objective of MDT is to leverage multiple labeled datasets $\mathcal{D}$ to train a unified model that achieves a more generalizable function $F: \mathcal{X} \rightarrow \mathcal{L}$. 
For MDT, a straightforward strategy is to merge all data into a significantly larger dataset $ \mathcal{D}_\text{total} = \mathcal{D}_1 \cup  \mathcal{D}_2 \cup \cdots$ and amalgamate their label spaces $\mathcal{L}_\text{total}=\mathcal{L}_1 \cup \mathcal{L}_2 \cup \cdots$.
This approach optimizes the same loss function over an expanded dataset:
\begin{equation}
    \min_{\Theta} \mathbb{E}_{(\hat X, \hat L) \in \mathcal{D}_\text{total}} \left[\ell(\mathcal{F}(\hat X; \Theta), \hat{L}_\text{total})\right]
\end{equation}



In the \cref{sec:framework,sec:GR,sec:SLM}, we present the design of the MergeOcc framework and show how to train a 3D perception model that performs well on multiple LiDARs.

\subsection{Overview of MergeOcc}
\label{sec:MergeOcc}

\subsubsection{Overview Framework}
\label{sec:framework}

The ambition of MergeOcc is to construct a comprehensive model capable of learning across $K$ distinct datasets $\mathcal{D}_1, \cdots, \mathcal{D}_K$ with respective label spaces $\mathcal{L}_1, \cdots, \mathcal{L}_K$
. 
Given the uniform task objective and similar data structure of the input, 
the network can share the backbone parameters and just use dataset-specific heads. 
This can be considered training
 $K$ partitioned dataset-specific models $\mathcal{F}_1, \cdots, \mathcal{F}_K$ in parallel, where each dataset uses its own classification layer and loss $\ell_k$ at the end.
The objective of the training process is to minimize the cumulative loss across all datasets, as formalized in the following equation:
\begin{equation}
    \min_{\Theta} \mathbb{E}_{D_k}\left[\mathbb{E}_{(\hat X, \hat L) \in D_k}\left[ \ell_k(\mathcal{F}_k(\hat X; \Theta), \hat L_\text{total})\right]\right]
\end{equation}

The MergeOcc framework is comprehensively illustrated in \cref{fig:framework}, showcasing the sequential processing stages from the acquisition of raw LiDAR data to the generation of fine-grained occupancy prediction outputs. 
The core components of MergeOcc comprise a geometric realignment module and a semantic label mapping module. The former involves point range alignment and dataset-specific normalization to ensure adaptability across different LiDARs. The latter is conducted on the output to transcode the original categories to a unified label space. 

Inspired by OpenOccupancy, MergeOcc incorporates an upsampling process for coarse-grained occupancy grids through geometric coarse-to-fine queries. While the primary objective of upsampling is to balance model performance with GPU memory usage, we posit that it serves to achieve self-regression outputs through a non-autoregressive mode, thereby enhancing the model's efficiency and effectiveness. The implementation is detailed in \cref{sec:appendixe}


\begin{table*}[!thb]
\small
\centering

{
\begin{tabular}{ccccc}
\hline
\multirow{2}{*}{Methods} & \multirow{2}{*}{OpenOccupancy-nuScenes Range} & \multirow{2}{*}{SemantickittiKITTI Range} & tested on OO-nu & tested on SK \\
 &  &  & $\text{geometric IoU}$ &  $\text{geometric IoU}$  \\ 
\hline
\multirow{3}{*}{Not Align.} & L=[-51.2, 51.2]m & L=[-72.0, 72.0]m & \multirow{3}{*}{25.4} & \multirow{3}{*}{36.7} \\
& W=[-51.2, 51.2]m & W=[-72.0, 72.0]m  & \\
& H=[-5.0, 3.0]m & H=[-3.4, 3.0]m & \\
\toprule[0.5pt]
\multirow{3}{*}{Align Max} & L=[-51.2, 51.2]m & L=[-51.2, 51.2]m & \multirow{3}{*}{16.5{($-$8.9)}} & \multirow{3}{*}{53.1{($+$16.4)}} \\
& W=[-51.2, 51.2]m & W=[-51.2, 51.2]m  & \\
& H=[-5.0, 3.0]m & H=[-5.0, 3.0]m & \\
\hline
\multirow{3}{*}{Align Min(MergeOcc)} & L=[0, 51.2]m & L=[0, 51.2]m & \multirow{3}{*}{\textbf{38.2}{($+$12.8)}} & \multirow{3}{*}{\textbf{70.0}{($+$33.3)}} \\
& W=[-25.6, 25.6]m & W=[-25.6, 25.6]m  & \\
& H=[-3.4, 3.0]m & H=[-3.4, 3.0]m & \\
\hline
\end{tabular}
}
\caption{Inconsistent LiDAR ranges will cause a significant model performance drop. For the intuitiveness of presentation, we only show the geometric IoU. The model employs L-CONet~\cite{wangOpenOccupancyLargeScale2023} trained on both datasets.}
\label{tab2}
\end{table*}

\subsection{Geometric Realignment}
\label{sec:GR}
\subsubsection{Geometric Disparities} 
Unlike 2D images, which are organized into a fixed grid structure of pixels, 3D point clouds are acquired through a set of lasers that varies across LiDARs, so inherently unordered. This variability introduces a geometric distributional discrepancy across datasets. The principal differences between the two datasets under study are detailed in ~\cref{tab1}. 
\subsubsection{Point Cloud Range Alignment} 
Drawing inspiration from Uni3D\cite{zhangUni3DUnifiedBaseline2023}, we investigate the impact of point cloud range on dense-predictive generative tasks. The results, delineated in \cref{tab2}, prompt us to implement distinctly different restriction strategies based on the task at hand.
 
For the object detection task, Uni3D advocates for utilizing the maximum point cloud range as a metric for correction. This approach is justified by the observation that objects are typically located within areas covered by the LiDAR point cloud returns. Consequently, selecting an extended range enhances the model's capability to detect objects at greater distances. However, our analysis, as summarized in ~\cref{tab2}, reveals that employing the maximum point cloud range for realignment in dense-predict generative tasks adversely affects performance. The discrepancy in occupancy grid shapes and annotation ranges will disrupt the learning process, leading to confusion in areas of annotation inconsistency for unified models. Therefore, opting for the minimum point cloud range for the occupancy prediction task yields more favorable outcomes than the maximum.

\subsubsection{Geometric Statistic Data Alignment} 
As a proven and effective approach in transfer learning~\cite{chattopadhyayPASTAProportionalAmplitude2023,zhangUni3DUnifiedBaseline2023}, a dataset-specific normalization operation is introduced to bridge the domain gap between different LiDARs. Our intuition is to encapsulate the unique statistical characteristics of each LiDAR within our model framework. The  geometric statistic data realignment strategy is delineated as ~\cref{eq:norm}:

\begin{equation}
    \label{eq:norm}
    \hat{y}_k^j \ = \ \gamma^j \hat{x}_k^j + \beta^j = \gamma^j \frac{x_k^j - \mu^j_k}{\sqrt{\sigma_k^j + \xi}} + \beta^j
\end{equation}

\noindent where the $\mu^j_k$ and $\sigma_k^j$ represent the dataset-specific mean and variance corresponding to the input samples, the $\xi$ is introduced for numerical stability, and the 
parameters $\gamma$ and $\beta$ are utilized to restore the model's representational capacity. 

Traditional BatchNorm~\cite{ioffeBatchNormalizationAccelerating2015}  estimates the statistical properties of the data within the current batch by standardizing the inputs for each layer across a network to attain a mean $\mu^j$ of $0$ and a standard deviation $\sigma^j$ of $1$. However, this conventional normalization relying on shared statistics does not make sense and may impact model performance in the MDT paradigm, where data within one batch may originate from different datasets. Therefore, we utilize dataset-specific mean and variance to encapsulate diverse statistical data. Notably, $\gamma$ and $\beta$ are shared across datasets. This strategy is justified by the alignment of inter-dataset discrepancies in first and second-order statistics post-normalization, thereby rationalizing the adoption of uniform $\gamma$ and $\beta$ parameters across diverse datasets.

The geometric realignment can be easily integrated with previous 3D perception models, enhancing their adaptability and performance across disparate data sources.

\subsection{Semantic Label Mapping (SLM)}
\label{sec:SLM}
\subsubsection{Motivation of SLM}
As delineated in \cref{sec:appendixf}, semantic discrepancies across different datasets, stemming from varying class definitions and annotation granularity,  present significant challenges in the MDT paradigm. 
Specifically, multi-head models often yield duplicate outputs for identical objects that appear in multiple datasets, leading to uncertainty and redundancy that adversely impact downstream tasks. Therefore, SLM is necessary for aligning initial outputs to a unified label space in the MDT paradigm.

Traditional approaches include manually crafted mappings $\mathcal{T}$~\cite{lambertMSegCompositeDataset2020,wangUniversalObjectDetection2019}, and LLM-based merging strategies ~\cite{caiBigDetectionLargescaleBenchmark2022}. However, these methods often encounter the hurdle of label ambiguity, necessitating considerable human effort for label correction. 
As the number of dataset categories increases, the complexity of these methods becomes unacceptable.
Additionally, both Wang et al.\citeyearpar{wangUniversalObjectDetection2019} and the MSeg~\cite{lambertMSegCompositeDataset2020} have reported obvious performance degradation of unified models compared to dataset-specific models, due to the complexities involved in integrating multiple domain sources.

To address these issues, we introduce SLM to obtain unique semantic outputs in an automated and scalable manner, while maintaining the generalized model's performance.

\begin{figure*}[!thb]
  \centering
  \begin{subfigure}{0.373\textwidth}
    \includegraphics[width=\linewidth]
    {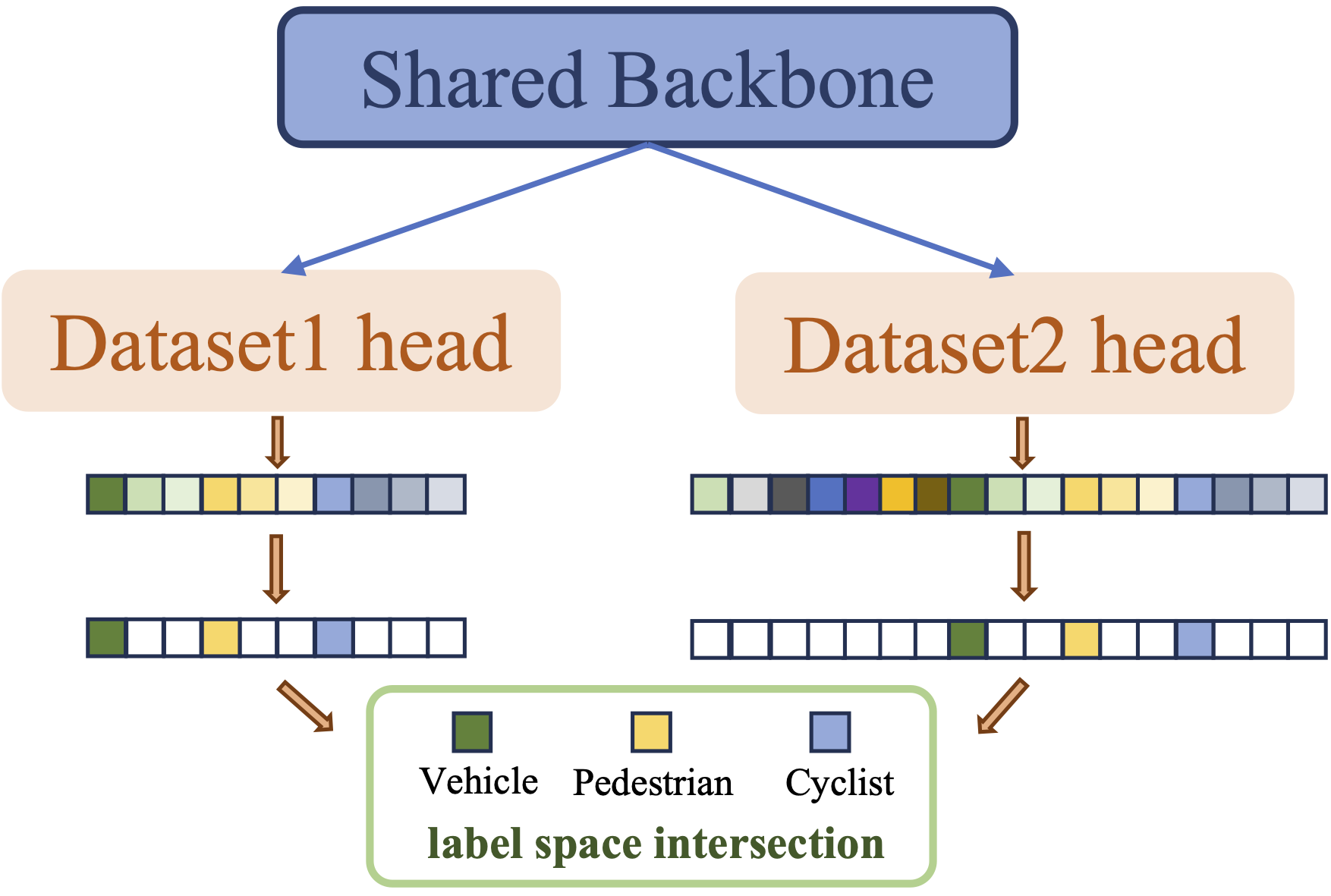}
    \centering
    \caption{ST3D et al. use the intersection}
    \label{fig:label_mapping1}
  \end{subfigure}
  \hfill
  \begin{subfigure}{0.622\textwidth}
    \includegraphics[width=\linewidth]{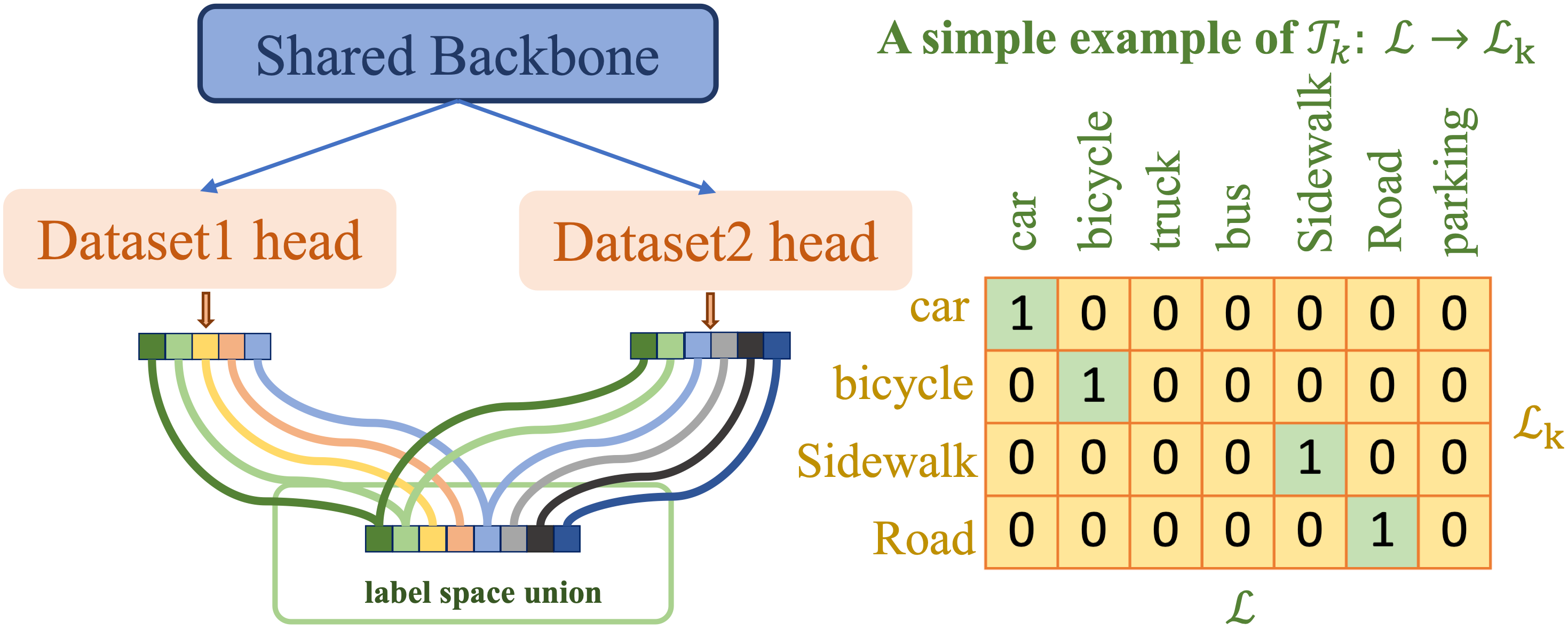}
    \centering
    \caption{MergeOcc use the union of label space}
    \label{fig:label_mapping2}
  \end{subfigure}
  \caption{Comparison of previous label space intersection and our unified label space}
  \label{fig:label_mapping}
\end{figure*}

\subsubsection{Learn a Unified Label Space}
In the realm of MDT, each dataset with its own label space $\mathcal{L}_1, \mathcal{L}_2, \cdots$, collectively ascertain a unified label space $\mathcal{L}$. This involves establishing a mapping function ${\mathcal{T}_k: \mathcal{L} \to \mathcal{L}_k}$ for each dataset, enabling the conversion of dataset-specific labels to a unified label space. Furthermore, the transformation ${\mathcal{T}_j \mathcal{T}_i^\top}$ facilitates the translation of output from label space $\mathcal{L}_i$ to another $\mathcal{L}_j$. 

Previous MDT-related 3D object detection studies~\cite{yangST3DDenoisedSelftraining2022, yangST3DSelftrainingUnsupervised2021, zhangUni3DUnifiedBaseline2023} predominantly employed a straightforward approach for label space merging, focusing solely on the critical intersection of label spaces, as illustrated in \cref{fig:label_mapping1}. 
This approach confines their attention to a limited subset of categories—namely, vehicles, pedestrians, and cyclists—owing to their importance in autonomous driving contexts. While this is a prevalent practice, it may impede the models' generalization to real-world scenarios. 
To address this limitation, inspired by Zhou et al.\citeyearpar{zhouSimpleMultidatasetDetection2022}, we introduce a data-driven optimization method for automatically constructing a unified label space encompassing a broader spectrum of categories, as depicted in \cref{fig:label_mapping2}. 
This union strategy enhances the model's generalization ability by integrating a more comprehensive and representative set of categories, aligning more closely with the complex and varied nature of the real world. 

Mathematically, $\mathcal{T}_k\in \{0,1\}^{|\mathcal{L}_k|\times |\mathcal{L}|}$ represents a boolean linear transformation.
Each label, $l_k \in \mathcal{L}_k$, specific to a dataset, is mapped to exactly one unified label $l_\text{uni} \in \mathcal{L}$ such that: $\mathcal{T}_k \vec 1 = \vec 1$. Furthermore, each unified label corresponds to at most one dataset-specific label, as indicated by: $\mathcal{T}_k^\top \vec 1 \leq \vec 1$, which implies that no dataset contains duplicate classes. Notably, these two constraints do not establish hierarchical relationships across datasets but preserve all granularities within the unified label space, accommodating varying label granularities. Our objective is to ascertain a set of mappings $\mathcal{T} = \left[\mathcal{T}_1, \cdots, \mathcal{T}_N\right]$ and implicitly define a unified label-space $\mathcal{L}$, maintaining the performance of the unified model unaffected.

Further analysis of $\mathcal{T}_k$ facilitates subsequent efficient optimization. Notably, the cardinality of unified label set $\mathcal{L}$ corresponds to the number of columns of $\mathcal{T}$.
Additionally, since we merge at most one label per dataset: $\mathcal{T}_k^{\top} \vec 1 \le \vec 1$.
Thus, for each dataset $\mathcal{D}_k$ a column $\mathcal{T}_k(c) \in \mathbb{T}_k$ takes one of $|\mathcal{\hat L}_k|+1$ values: $\mathbb{T}_k = \{\vec 0, \vec 1_{1}, \vec 1_{2}, \cdots\}$, where $\vec 1_{i} \in \{0,1\}^{|\mathcal{L}_k|}$ serves as an indicator vector of the $i$-th element. A simple example of $\mathcal{T}_k$ is illustrated in \cref{fig:label_mapping}.

Consequently, each column $\mathcal{T}(c) \in \mathbb{T}$ is selected from a limited set of potential values $\mathbb{T} = \mathbb{T}_1 \times \mathbb{T}_2 \times \cdots$, where  $\times$ representing the Cartesian product.
Instead of optimizing directly over the label set $\mathcal{L}$ and transformation $\mathcal{T}$, we adopt combinatorial optimization over the potential column values $\vec t \in \mathbb{T}$. 
Let $x_{\vec t} \in \{0,1\}$ denote whether combination $\vec t \in \mathbb{T}$ is selected.
In this formulation, the constraint $\mathcal{T}_k \vec 1 = \vec 1, \forall \ k \in N$ translates to $\sum_{\vec t\in\mathbb{T} \mid \vec t(c) = 1}x_{\vec t} = 1$ for all dataset-specific labels $c$. This scalable and automatic approach minimizes the need for extra human intervention for label rectification. 

In practice, we utilize correlations and similarity within the outputs of partitioned occupancy prediction models across various LiDAR datasets, which serve as proxies for perceptual consistency, thereby addressing the challenge of label ambiguity. 
For a set of dataset-specific models' occupancy predictions ${o_i^1 \in \mathbb{R}^{|\mathcal{L}_1|}, o_i^2 \in \mathbb{R}^{|\mathcal{L}_2|}, \cdots}$, we construct a joint prediction score $d_i$ by averaging the outputs on classes:
\begin{equation}
 d_i = \frac{\sum_k \mathcal{T}_k^\top \vec o^k}{\sum_k \mathcal{T}_k^\top \vec 1}
\end{equation}
where the division is elementwise.
The dataset-specific outputs can be restored by $\tilde o_i^k = \mathcal{T}_k d_i$.
Specifically, we seek to optimize the label mapping function by maximizing the similarity of the results obtained through two reciprocal mappings ${\mathcal{T}_k \mathcal{T}_k^\top o_i^k}$: from a specific dataset to the learned unified label space and then back to the original label space.

\definecolor{LightGrey}{rgb}{.9,.9,.9}
\definecolor{White}{rgb}{1.,0.,1.}
\definecolor{first}{rgb}{.8,.0,.0}
\definecolor{second}{rgb}{.0,.6,.0}
\definecolor{third}{rgb}{.0,.0,.8}

\definecolor{nbarrier}{RGB}{255, 120, 50}
\definecolor{nbicycle}{RGB}{255, 192, 203}
\definecolor{nbus}{RGB}{255, 255, 0}
\definecolor{ncar}{RGB}{0, 150, 245}
\definecolor{nconstruct}{RGB}{0, 255, 255}
\definecolor{nmotor}{RGB}{200, 180, 0}
\definecolor{npedestrian}{RGB}{255, 0, 0}
\definecolor{ntraffic}{RGB}{255, 240, 150}
\definecolor{ntrailer}{RGB}{135, 60, 0}
\definecolor{ntruck}{RGB}{160, 32, 240}
\definecolor{ndriveable}{RGB}{255, 0, 255}
\definecolor{nother}{RGB}{139, 137, 137}
\definecolor{nsidewalk}{RGB}{75, 0, 75}
\definecolor{nterrain}{RGB}{150, 240, 80}
\definecolor{nmanmade}{RGB}{213, 213, 213}
\definecolor{nvegetation}{RGB}{0, 175, 0}

\definecolor{car}{rgb}{0.39215686, 0.58823529, 0.96078431}
\definecolor{bicycle}{rgb}{0.39215686, 0.90196078, 0.96078431}
\definecolor{motorcycle}{rgb}{0.11764706, 0.23529412, 0.58823529}
\definecolor{truck}{rgb}{0.31372549, 0.11764706, 0.70588235}
\definecolor{other-vehicle}{rgb}{0.39215686, 0.31372549, 0.98039216}
\definecolor{person}{rgb}{1.        , 0.11764706, 0.11764706}
\definecolor{bicyclist}{rgb}{1.        , 0.15686275, 0.78431373}
\definecolor{motorcyclist}{rgb}{0.58823529, 0.11764706, 0.35294118}
\definecolor{road}{rgb}{1.        , 0.        , 1.        }
\definecolor{parking}{rgb}{1.        , 0.58823529, 1.        }
\definecolor{sidewalk}{rgb}{0.29411765, 0.        , 0.29411765}
\definecolor{other-ground}{rgb}{0.68627451, 0.        , 0.29411765}
\definecolor{building}{rgb}{1.        , 0.78431373, 0.        }
\definecolor{fence}{rgb}{1.        , 0.47058824, 0.19607843}
\definecolor{vegetation}{rgb}{0.        , 0.68627451, 0.        }
\definecolor{trunk}{rgb}{0.52941176, 0.23529412, 0.        }
\definecolor{terrain}{rgb}{0.58823529, 0.94117647, 0.31372549}
\definecolor{pole}{rgb}{1.        , 0.94117647, 0.58823529}
\definecolor{traffic-sign}{rgb}{1.        , 0.        , 0.    }   

\makeatletter
\newcommand{\car@semkitfreq}{3.92}
\newcommand{\bicycle@semkitfreq}{0.03}
\newcommand{\motorcycle@semkitfreq}{0.03}
\newcommand{\truck@semkitfreq}{0.16}
\newcommand{\othervehicle@semkitfreq}{0.20}
\newcommand{\person@semkitfreq}{0.07}
\newcommand{\bicyclist@semkitfreq}{0.07}
\newcommand{\motorcyclist@semkitfreq}{0.05}
\newcommand{\road@semkitfreq}{15.30}  %
\newcommand{\parking@semkitfreq}{1.12}
\newcommand{\sidewalk@semkitfreq}{11.13}  %
\newcommand{\otherground@semkitfreq}{0.56}
\newcommand{\building@semkitfreq}{14.1}  %
\newcommand{\fence@semkitfreq}{3.90}
\newcommand{\vegetation@semkitfreq}{39.3}  %
\newcommand{\trunk@semkitfreq}{0.51}
\newcommand{\terrain@semkitfreq}{9.17} %
\newcommand{\pole@semkitfreq}{0.29}
\newcommand{\trafficsign@semkitfreq}{0.08}
\newcommand{\semkitfreq}[1]{{\csname #1@semkitfreq\endcsname}}

\newcommand{\barrier@nuscenesfreq}{11.79}
\newcommand{\bicycle@nuscenesfreq}{0.18}
\newcommand{\bus@nuscenesfreq}{5.83}
\newcommand{\car@nuscenesfreq}{48.27}
\newcommand{\construction@nuscenesfreq}{1.92}
\newcommand{\motorcycle@nuscenesfreq}{0.54}
\newcommand{\pedestrian@nuscenesfreq}{2.93}
\newcommand{\trafficcone@nuscenesfreq}{0.93}
\newcommand{\trailer@nuscenesfreq}{6.22}
\newcommand{\truck@nuscenesfreq}{20.07}
\newcommand{\driveable@nuscenesfreq}{28.64}
\newcommand{\other@nuscenesfreq}{0.77}
\newcommand{\sidewalk@nuscenesfreq}{6.34}
\newcommand{\terrain@nuscenesfreq}{6.35}
\newcommand{\manmade@nuscenesfreq}{16.10}
\newcommand{\vegetation@nuscenesfreq}{11.08}
\newcommand{\nuscenesfreq}[1]{{\csname #1@nuscenesfreq\endcsname}}

\begin{table*}[!hbt]
\footnotesize
\setlength{\tabcolsep}{0.008\linewidth}
\centering
\setlength{\tabcolsep}{0.9mm}
{
\begin{tabular}{c | c | c c | c c c c c c c c c c c c c c c c}
\toprule
Method & Input & IoU & mIoU
& \rotatebox{90}{\textcolor{nbarrier}{$\blacksquare$} barrier}
& \rotatebox{90}{\textcolor{nbicycle}{$\blacksquare$} bicycle}
& \rotatebox{90}{\textcolor{nbus}{$\blacksquare$} bus}
& \rotatebox{90}{\textcolor{ncar}{$\blacksquare$} car}
& \rotatebox{90}{\textcolor{nconstruct}{$\blacksquare$} const. veh.}
& \rotatebox{90}{\textcolor{nmotor}{$\blacksquare$} motorcycle}
& \rotatebox{90}{\textcolor{npedestrian}{$\blacksquare$} pedestrian}
& \rotatebox{90}{\textcolor{ntraffic}{$\blacksquare$} traffic cone}
& \rotatebox{90}{\textcolor{ntrailer}{$\blacksquare$} trailer}
& \rotatebox{90}{\textcolor{ntruck}{$\blacksquare$} truck}
& \rotatebox{90}{\textcolor{ndriveable}{$\blacksquare$} drive. suf.}
& \rotatebox{90}{\textcolor{nother}{$\blacksquare$} other flat}
& \rotatebox{90}{\textcolor{nsidewalk}{$\blacksquare$} sidewalk}
& \rotatebox{90}{\textcolor{nterrain}{$\blacksquare$} terrain}
& \rotatebox{90}{\textcolor{nmanmade}{$\blacksquare$} manmade}
& \rotatebox{90}{\textcolor{nvegetation}{$\blacksquare$} vegetation}
\\
\midrule
MonoScene~\citeyearpar{caoMonoSceneMonocular3D2022} & C & 18.4 & 6.9 & 7.1 & 3.9 & 9.3 & 7.2 & 5.6 & 3.0 & 5.9 & 4.4 & 4.9 & 4.2 & 14.9 & 6.3 & 7.9 & 7.4 & 10.0 & 7.6 \\

TPVFormer~\citeyearpar{huangTriPerspectiveViewVisionBased2023} & C & 15.3 & 7.8 & 9.3 & 4.1 & 11.3 & 10.1 & 5.2 & 4.3 & 5.9 & 5.3 & 6.8 & 6.5 & 13.6 & 9.0 & 8.3 & 8.0 & 9.2 & 8.2 \\

3DSketch~\citeyearpar{chen3DSketchawareSemantic2020} & \makecell{C\&D} & 25.6 & 10.7 & 12.0 & 5.1 & 10.7 & 12.4 & 6.5 & 4.0 & 5.0 & 6.3 & 8.0 & 7.2 & 21.8 & 14.8 & 13.0 & 11.8 & 12.0 & 21.2 \\

LMSCNet~\citeyearpar{roldaoLMSCNetLightweightMultiscale2020} & L & 27.3 & 11.5 & 12.4 & 4.2 & 12.8 & 12.1 & 6.2 & 4.7 & 6.2 & 6.3 & 8.8 & 7.2 & 24.2 & 12.3 & 16.6 & 14.1 & 13.9 & 22.2 \\

JS3C-Net~\citeyearpar{yanSparseSingleSweep2021} & L & 30.2 & 12.5 & 14.2 & 3.4 & 13.6 & 12.0 & 7.2 & 4.3 & 7.3 & 6.8 & 9.2 & 9.1 & 27.9 & 15.3 & 14.9 & 16.2 & 14.0 & 24.9 \\

C-CONet~\citeyearpar{wangOpenOccupancyLargeScale2023} & C & 20.1 & 12.8 & 13.2 & 8.1 & 15.4 & 17.2 & 6.3 & 11.2 & 10.0 & 8.3 & 4.7 & 12.1 & 31.4 & 18.8 & 18.7 & 16.3 & 4.8 & 8.2  \\

L-CONet~\citeyearpar{wangOpenOccupancyLargeScale2023} & L & 30.9 & 15.8 & 17.5 & 5.2 & 13.3 & 18.1 & 7.8 & 5.4 & 9.6 & 5.6 & 13.2 & 13.6 & 34.9 & {21.5} & 22.4 & 21.7 & 19.2 & 23.5  \\

M-CONet~\citeyearpar{wangOpenOccupancyLargeScale2023} & \makecell{C\&L} & 29.5 & 20.1 & {23.3} & 13.3 & {21.2} & 24.3 & \textbf{15.3} & \underline{15.9} & 18.0 & \underline{13.3} & \underline{15.3} & {20.7} & 33.2 & 21.0 & 22.5 & 21.5 & 19.6 & 23.2   \\

SurroundOcc~\citeyearpar{weiSurroundOccMultiCamera3D2023} & C & {31.5} & {20.3}  & {20.6} & \underline{11.7} & \textbf{28.0} & \textbf{30.9} & {10.7} & {15.1} & {14.1} & {12.1} & {14.4} & \textbf{22.3} & \underline{37.3} & {23.7} & {24.5} & 22.8 & {14.9} & {21.9} \\

PointOcc ~\citeyearpar{zuoPointOccCylindricalTriperspective2023} & L & \underline{{34.1}} & \textbf{{23.9}} & \textbf{24.9} & \textbf{19.0} & {20.9}& {25.7} & \underline{13.4} & \textbf{25.6} & \textbf{30.6} & \textbf{17.9} & \textbf{16.7} & \underline{21.2} & {36.5} & \textbf{25.6} & \underline{25.7} & \underline{24.9} & \underline{24.8} & \underline{29.0} \\ 

\midrule



MergeOcc (D.M.)(ours) & L & {25.4} & {5.7} & {2.0} & {0.1} & 1.4 & {5.4} & 0.3 & {0.3} & {0.4} & {0.3} & {2.8} & {2.5} & {28.1} & {10.3} & {10.9} & {10.3} & {6.5} & {9.9}  \\ 

MergeOcc (Tr.both)(ours) & L & \textbf{{40.2}} & \underline{{22.9}} & \underline{23.4} & {8.8} & \underline{22.9} & \underline{27.5} & \underline{13.4} & {15.8} & \underline{21.9} & {12.7} & \underline{15.3} & {19.4} & \textbf{39.8} & \underline{25.0} & \textbf{26.4} & \textbf{27.5} & \textbf{30.8} & \textbf{35.0}  \\ 

\bottomrule
\end{tabular}
}

\caption{\textbf{3D Semantic occupancy prediction results on OpenOccupancy-nuScenes.} We report the performance on semantic scene completion (SSC - mIoU) and geometric scene completion (SC - IoU) for LiDAR-inferred baselines and our method. The C, L, and D denotes camera, LiDAR, and depth, respectively. }
\label{tab: nu occ result}
\end{table*}

Consider $O^k = [o_1^k, o_2^k, \cdots]$ as the outputs of the dataset-specific occupancy head for dataset $\mathcal{D}_k$, let $\ell_c$ denote a loss function that assesses the fidelity of the transcoded results $O_\text{uni}^k$ with unified label $l_\text{uni} \in \mathcal{L}$ and its re-projections $\hat O^k$ with original label $l_i^k \in \mathcal{L}_k$ in the occupancy grids.

We define $D = \frac{\sum_k \mathcal{T}_k^\top O^k}{\sum_k \mathcal{T}_k^\top \vec 1}$ as the merged occupancy prediction scores, and $\tilde O^k = \mathcal{T}_k D$ as the re-projection. Building upon our earlier analysis of the nature of $\mathcal{T}_k$, our objective is to minimize this loss across all dataset-specific outputs while respecting the boolean constraints imposed on our mapping:
\begin{equation}
\mathcal{J} = \sum_{\vec t \in \mathbb{T}} x_{\vec t} 
\underbrace{E_{\mathcal{D}_k}\left[\sum_{c\in \mathcal{L}_k \mid \vec t(c) = 1}\mathcal{L}_c(O^k, \tilde O^k)\right]}_{c_{\vec t}} + \lambda \sum_{\vec t \in \mathbb{T}} x_{\vec t}
\label{objective}
\end{equation}
Here, the penalty term $\lambda \sum_{\vec t \in \mathbb{T}} x_{\vec t}$ is introduced to promote a small and concise label space. The merge cost $c_{\vec t}$ can be precomputed for any subset of labels $\vec t$ facilitating a computationally efficient approach.
This yields a concise integer linear programming formulation of the objective~\cref{objective}:
\begin{align}
&\text{minimize}_{x} & \mathcal{J} = \sum_{\vec t \in \mathbb{T}}  & x_{\vec t} \left(c_{\vec t} + \lambda\right) \nonumber\\
&\text{subject to} & \sum_{\vec t\in\mathbb{T} \mid \vec t_{c} = 1}&x_{\vec t} = 1 \qquad \forall {c} \in \mathcal{L}_k &
\end{align}

For the scenario involving two datasets, this objective corresponds to weighted bipartite matching.
For a higher number of datasets, it reduces to weighted graph matching and is NP-hard yet practically solvable with integer linear programming. One limitation of this combinatorial approach is the exponential growth of the set of potential combinations $\mathbb{T}$ with the number of datasets employed: $|\mathbb{T}| = O(|\hat L_1| |\hat L_2| |\hat L_3| \ldots)$. A sequential paradigm, detailed in the \cref{sec:appendixf}, can be adopted to mitigate this complexity.

\section{Experiments}
\label{sec:Exp}
Detailed experimental settings and supplementary results are provided in the \cref{sec:appendixb,sec:appendixc,sec:appendixd}. In this section, we primarily present key experimental findings and discuss the underlying reasons for these outcomes.

\begin{table*}[!htb]
\centering
\begin{small}
\setlength{\tabcolsep}{0.01\linewidth}
{
    \begin{tabular}{c|c|c|c|c|c}
        \bottomrule[1pt]
        \multirow{2}{*}{Trained on} & \multirow{2}{*}{Method}  & \multicolumn{2}{c|}{Tested on OpenOccupancy-nuScenes} & \multicolumn{2}{c}{Tested on SemanticKITTI} \\
        \cline{3-6}
        &  & Geometric IoU & Semantic mIoU & Geometric IoU & Semantic mIoU \\
        \hline
        \multirow{2}{*}{only SK} & MergeOcc  (w/o \texttt{P.T.})   & 4.1  &  0.1  &  55.6  & 4.7  \\
        & MergeOcc  (w/ \texttt{P.T.} on OO-nu)  & 4.3{($+$0.2)}  & 0.1   & 53.3{($-$2.3)}  & 4.0{($-$0.7)}  \\
        \toprule[1pt]
        \multirow{2}{*}{only OO-nu} & MergeOcc  (w/o \texttt{P.T.})  & 30.1  &  15.8  & 3.8  & 0.3  \\
        & MergeOcc  (w/ \texttt{P.T.} on SK)  & 26.3{($-$3.8)}  & 7.1{($-$8.7)}  & 5.7{($+$1.9)}  & 0.1{($-$0.2)}  \\
        \toprule[1pt]
        \multirow{5}{*}{SK+OO-nu} & MergeOcc  (w/ \texttt{D.M.})   &  25.4  & 5.7  & 36.7 & 2.0 \\
        & MergeOcc  (w/ \texttt{G.A.})  &  14.7{($-$10.7)}  & 1.6{($-$4.1)}  &  51.2{($+$14.5)}  &  3.0{($+$1.0)}  \\
        & MergeOcc  (w/ \texttt{R.A.})  & {38.2}{($+$12.8)}  & {21.3}{($+$15.6)}  & \underline{70.0}{($+$33.3)}  & \underline{13.4}{($+$11.4)}   \\
        & MergeOcc  (w/ \texttt{G.R.(R.A.+G.A.)})  & \underline{38.4}{($+$13.0)}  & \underline{21.4}{($+$15.7)}  & {69.9}{($+$33.2)}  &  \textbf{14.5}{($+$12.5)}  \\
        & MergeOcc  (w/ \texttt{G.R.+ SLM})  & \textbf{40.2}{($+$14.8)}  & \textbf{22.9}{($+$17.2)}  & \textbf{70.1}{($+$33.4)}  &  \textbf{14.5}{($+$12.5)}  \\
        \toprule[1pt]
    \end{tabular}
}
\caption{Ablation results of core components of MergeOcc on OpenOccupancy-nuScenes and SemanticKITTI datasets.}
\label{tab: ablation}
\end{small}
\vspace{-0.2cm}
\end{table*}

\subsection{3D Semantic Occupancy Prediction Results}
As illustrated in~\cref{tab: nu occ result} and~\cref{tab:semkitti result}(\cref{sec:appendixc}), the experimental results underscore the superior performance and robust cross-LiDAR scalability of our approach.

\noindent\textbf{Compared with D.M. Method}
MergeOcc effectively bridges the domain gap between disparate LiDARs, enabling the model to leverage broader and more diverse data.
Specifically, without bells and whistles, in the OpenOccupancy-nuScenes, MergeOcc enhances occupancy prediction geometric IoU and semantic mIoU by \textbf{14.8\%} and \textbf{17.2\%}. In the SemanticKITTI, MergeOcc improves the occupancy prediction accuracy by \textbf{33.40\%} and \textbf{12.50\%} respectively, marking a significant performance advancement.

\noindent\textbf{Compared with Baseline Trained on Single Domain}
A crucial and intriguing observation is that, despite using an identical architecture and hyper-parameter set, MergeOcc demonstrates a notable performance enhancement due to the exposure to broader data, which is unprecedented in the field of 3D perception. Specifically, the in-domian performance improvements are as follows: ${G}_\text{nu}:\mathbf{+9.3\%}, {S}_\text{nu}:\mathbf{+7.1\%}$; ${G}_\text{sk}:\mathbf{+14.5\%}, {S}_\text{sk}:\mathbf{+9.8\%}$. This substantial gain indicates that the \textbf{current capacity of 3D Occ datasets is insufficient} and highlights the crucial potential for the MDT paradigm.

Notably, MergeOcc simultaneously exhibits excellent performance on both types of LiDARs. When evaluated under the cross-LiDAR setting, MergeOcc's performance significantly surpasses the dataset-specific baseline.

\noindent\textbf{Compared with SOTA}
Moreover, MergeOcc outperforms LiDAR-based SOTA method PointOcc\cite{zuoPointOccCylindricalTriperspective2023} by $\mathbf{6.1\%}$ in IoU. Additionally, it demonstrates superior performance relative to the SOTA multi-modality model M-CONet by $\mathbf{10.7\%}$ and $\mathbf{2.8\%}$ in $\mathbf{IoU}$ and $\mathbf{mIoU}$, underscoring its efficacy in extracting insights from diverse data.

\begin{figure}[!htb]
\centering
\includegraphics[width=\linewidth]{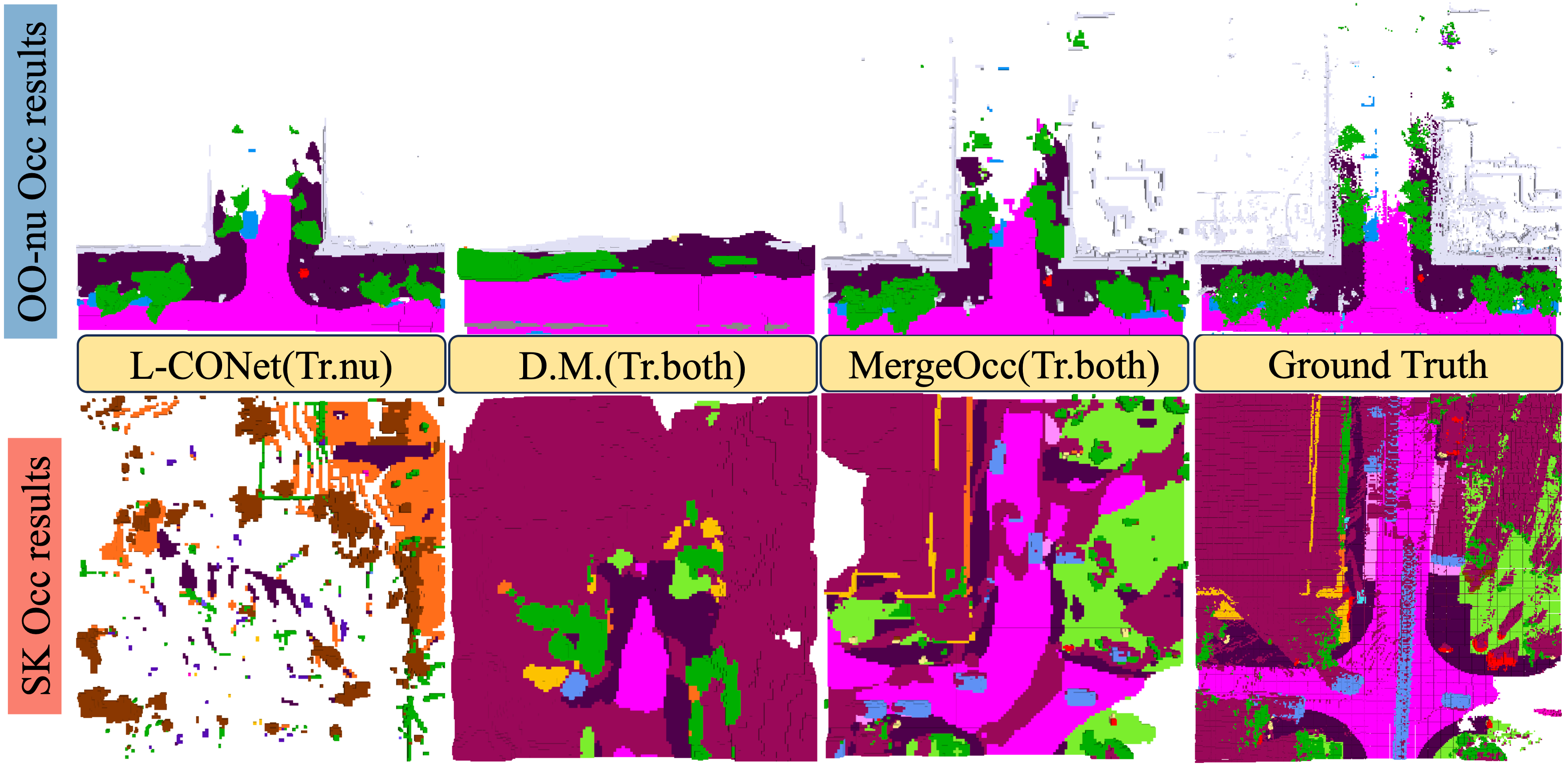}
\caption{Visualizations of occupancy prediction results. The upper half illustrates the outcomes of nuScenes, while the lower half showcases the SemanticKITTI results   .}
\vspace{-0.2cm}
\label{fig: visualization}
\end{figure}

The visual comparison of occupancy prediction results generated by the primary methods is illustrated in~\cref{fig: visualization}. Additional visualizations can be found in the \cref{sec:appendixh}.

\subsection{Ablation Experiments}

\subsubsection{Ablation Design and Comparison of Methods}
\noindent1) w/o \texttt{P.T.(Single-dataset)}: We utilize the off-the-shelf occupancy prediction network L-CONet~\cite{wangOpenOccupancyLargeScale2023} as baseline models, which is trained \underline{from scratch} and evaluated within a single dataset.

\noindent2) \texttt{P.T.(Pre-training)}:  We adopt the standard \underline{fine-tune paradigm}, where the baseline model is initially pre-trained on the source domain and subsequently fine-tuned on the target domain. The training processes for the two datasets are conducted independently.

\noindent3) \texttt{D.M.(Direct Merging)}: By simply combing multiple 3D datasets into a single merged dataset, dataset-specific models can be trained on it with a combined loss, which serves as a direct method to assess whether the existing 3D models can be improved by the directly-merged datasets. 

\noindent4) \texttt{G.R.(R.A.+G.A.)}: As introduced above, we perform point cloud range alignment(R.A.) and geometric statistical data alignment(G.A.) to bridge the domain gap between different LiDARs and encapsulate the unique statistical characteristics of each LiDAR into our model framework.

\noindent5) \texttt{G.R.+SLM}: This represents the complete MergeOcc. 

Notably, for all settings, we use an identical hyper-parameter set and training strategy to ensure fairness.

\subsubsection{Further Analysis}
\noindent\textbf{1) Large disparity across 3D LiDAR datasets}: As illustrated in ~\cref{tab: ablation}, dataset-specific models demonstrate adequate performance solely within their corresponding domains. However, their efficacy significantly deteriorates when applied to unseen domains. This cross-LiDAR issue primarily arises from the over-fitting to the source domains and unreasonable shortcuts of dataset-specific models, thereby overlooking the inherent shift between source and target domains.

\noindent\textbf{2) Limited efficacy of pre-training strategy and its detrimental impact on the target domain}: Adhering to the prevailing fine-tuning paradigm, we pre-train the baseline model on the source domain and subsequently fine-tune it on the target domain. Upon comparing the baselines \texttt{P.T} and w/o~\texttt{P.T.}, as depicted in ~\cref{tab: ablation}, it is evident that despite pre-training, instances of catastrophic forgetting persist, e.g. the model forgets knowledge acquired from the preceding pre-trained dataset. Although there is a marginal enhancement within the source domain, the pre-existing knowledge hinders the models' performance in the target domain.

\noindent\textbf{3) Assessment of MergeOcc's effectiveness and robustness}: By comparing w/ \texttt{P.T.} and \texttt{D.M} baselines from ~\cref{tab: ablation}, it reveals that 3D single-dataset training paradigm fails to yield a robust and generalized perception model. As shown in \cref{tab: nu occ result,tab: ablation,tab:semkitti result}, MergeOcc consistently outperforms all baseline methods in IoU and mIoU, thus confirming its superior ability to bridge domain gaps across varied LiDAR systems and to fully exploit the available data.

Specifically, the \texttt{R.A.}, \texttt{G.A.}, and \texttt{SLM} modules all enhance the accuracy of occupancy prediction. It is critical to note that these improvements actually derive from the learning process from larger and more diverse datasets (i.e., MDT). These modules serve merely as the bridge to address the gap between different domains. Further ablation analyses are detailed in \cref{sec:appendixc,sec:appendixd}.

\section{Conclusion}
This paper pioneered the exploration of training strategies for constructing a generalized 3D occupancy prediction model capable of handling different LiDARs, yielding robust and effective perception models. The proposed MergeOcc, which integrates geometric realignment and semantic label mapping, successfully tackled challenges stemming from domain divergence, as validated by extensive experimentation on prominent benchmarks. The current approach does not yet extend to handling unseen LiDARs, we will endeavor to resolve this limitation in the future. 

\bibliography{Formatting-Instructions-LaTeX-2025}

\clearpage
\appendix

\section{Appendix}
\label{sec:appendix}
\subsection{Simplified Code}
\label{sec:appendixa}
\label{sec:code}
We provide a simplified version of the codebase for assessing the efficacy of MergeOcc, which is accessible anonymously in the supplementary material.
By strictly adhering to the guidelines provided in the Readme.md file, users can train a generalized occupancy prediction model showcasing outstanding performance on both the SemanticKITTI dataset~\cite{behleySemanticKITTIDatasetSemantic2019} and the OpenOccupancy-nuScenes dataset~\cite{wangOpenOccupancyLargeScale2023}.

A more comprehensive version of the codebase will be released in the near future.

\subsection{Experiments Settings}
\label{sec:appendixb}

\subsubsection{Dataset Selection}
Given the relative scarcity of large-scale outdoor occupancy datasets, those that are available predominantly consist of data collected using 32-beam and 64-beam LiDARs. In this context, selecting representative and predominant datasets is a judicious strategy. Specifically, the nuScenes dataset, which employs a 32-beam LiDAR, and the SemanticKITTI dataset, which utilizes a 64-beam LiDAR, represent the primary configurations in autonomous vehicles. Moreover, these two datasets are mostly used dataset for autonomous driving perception tasks. We also desire to evaluate the efficacy of our proposed method should alternative high-quality outdoor Occupancy datasets featuring different LiDAR become available in the future.

\subsubsection{Task Description and Benchmark}
The 3D semantic occupancy prediction has garnered significant attention for autonomous driving, necessitating the assignment of semantic labels to all regions within the spatial domain. Our evaluation is conducted on two widely recognized datasets for autonomous driving occupancy prediction: SemanticKITTI and OpenOccupancy-nuScenes. In SemanticKITTI, the dataset's perceptive field spans the range from $[-72.0m, -72.0m, -3.4m]$ to $[72.0m, 72.0m, 3m]$. However, the annotated region is restricted from $[0m, -25.6m, -3.4m]$ to $[51.2m, 25.6m, 3m]$, featuring a voxel resolution of $0.2m$, thereby yielding in a volumetric representation of $256 \times 256 \times 32$ voxels for occupancy prediction. Conversely, OpenOccupancy-nuScenes extends its perceptual range from $[-51.2m, -51.2m, -5m]$ to $[51.2m, 51.2m, 3m]$, maintaining an identical voxel resolution of $0.2m$, resulting in a volumetric grid of $512 \times 512 \times 40$ voxels. Our analysis is confined to the intersection of ground truth ranges, as delineated in \cref{sec:GR}, specifically from $[0m, -25.6m, -3.4m]$ to $[51.2m, 25.6m, 3m]$. 

\subsubsection{The Metric of Occupancy Prediction}
For 3D semantic occupancy prediction, we use the intersection over the union ($\mathbf{IoU}$) of occupied voxels, ignoring their
semantic class as the evaluation metric of the scene completion (SC) task and the $\mathbf{mIoU}$ of all semantic classes for the semantic scene completion (SSC) task.
	\begin{equation}
		\begin{aligned}
			\text{IoU} = &\frac{TP}{TP + FP + FN} \\
			\text{mIoU} = \frac{1}{C} & \sum_{i=1}^{C} \frac{TP_{i}}{TP_{i} + FP_{i} + FN_{i}}
		\end{aligned}
	\end{equation}
where $TP$, $FP$, $FN$ indicate the number of true positive, false positive, and false negative predictions. $C$ is the class number.

\subsubsection{Model Architecture}
Our methodology adopts the architectural framework utilized by L-CONet\cite{wangOpenOccupancyLargeScale2023} across both datasets. The input points are configured to perform cylindrical partitioning with dimensions $(\mathcal{H}_\text{in}, \mathcal{W}_\text{in}, \mathcal{D}_\text{in})=(512, 360, 32)$, representing radius, angle, and height, respectively. Within the encoder, we employ prevailing Resnet3D~\cite{heDeepResidualLearning2016} as the 3D backbone, augmented by an FPN~\cite{linFeaturePyramidNetworks2017} to integrate features across multiple scales. The occupancy head produces a voxel representation of dimensions $(\mathcal{H}_\text{out}, \mathcal{W}_\text{out}, \mathcal{D}_\text{out}) = (128, 90, 8)$. To enhance performance, we adopt a coarse-to-fine query strategy to upsample the output by a factor of $s=4$, following the methodology of L-CONet.

\subsubsection{Optimization}
Notably, throughout all our experiments, we retained the hyper-parameters from the baseline work OpenOccupancy without modification. This decision partially demonstrates the superior performance of our method.

During training on both datasets, we employ the Adam optimizer complemented by a weight decay of 0.01. A cosine learning rate scheduler initiates with a peak value of $3e-4$, accompanied by a linear warm-up phase for the initial 500 iterations. The occupancy prediction leverages a combination of classic cross-entropy loss, Lovasz-softmax loss~\cite{bermanLovaszSoftmaxLossTractable2018}, and an affinity loss to concurrently optimize the geometry and semantic metrics~\cite{caoMonoSceneMonocular3D2022}.

\subsubsection{Detailed Setting} 
All experimental procedures were executed using the mmdetection3d framework. Notably, disparities in point cloud range significantly impacted cross-dataset detection performance, as outlined in ~\cref{tab2}. To mitigate this issue, we standardized the point cloud ranges across all datasets to $[0, 51.2]m$ along the $X$ axis, $[-25.6, 25.6]m$ along the $Y$ axis, and $[-3.4, 3]m$ along the $Z$ axis, aligning with the intersecting range of the datasets under consideration. All models undergo training for 24 epochs, with a batch size of 8, distributed across 8 RTX 3090 GPUs. For the OpenOccupancy-nuScenes benchmark, we utilize multiple (10) LiDAR sweeps as input, adhering to a widely accepted practice. Conversely, for the SemanticKITTI dataset, a single LiDAR sweep is utilized as input.

Moreover, to ensure gradient backward propagation through all network parameters and sustain the training process, we design a balanced distributed group sampler that integrates data from distinct datasets in each batch. To the best of our knowledge, MergeOcc pioneers the application of the scene completion task on mmdetection3d and explores the MDT paradigm within this framework.

\begin{table*}[!hbt]
\scriptsize
\newcommand{\classfreq}[1]{{~\tiny(\semkitfreq{#1}\%)}}  %
\centering
\setlength{\tabcolsep}{1.2mm}
{
\begin{tabular}{l|c|c c|c c c c c c c c c c c c c c c c c c c}
        \toprule
        Method
        & Input
        & {IoU}  &{mIoU}
        & \rotatebox{90}{\textcolor{road}{$\blacksquare$} road\classfreq{road}} 
        & \rotatebox{90}{\textcolor{sidewalk}{$\blacksquare$} sidewalk\classfreq{sidewalk}}
        & \rotatebox{90}{\textcolor{parking}{$\blacksquare$} parking\classfreq{parking}} 
        & \rotatebox{90}{\textcolor{other-ground}{$\blacksquare$} other-grnd\classfreq{otherground}} 
        & \rotatebox{90}{\textcolor{building}{$\blacksquare$} building\classfreq{building}} 
        & \rotatebox{90}{\textcolor{car}{$\blacksquare$} car\classfreq{car}} 
        & \rotatebox{90}{\textcolor{truck}{$\blacksquare$} truck\classfreq{truck}} 
        & \rotatebox{90}{\textcolor{bicycle}{$\blacksquare$} bicycle\classfreq{bicycle}} 
        & \rotatebox{90}{\textcolor{motorcycle}{$\blacksquare$} motorcycle\classfreq{motorcycle}} 
        & \rotatebox{90}{\textcolor{other-vehicle}{$\blacksquare$} other-veh.\classfreq{othervehicle}} 
        & \rotatebox{90}{\textcolor{vegetation}{$\blacksquare$} vegetation\classfreq{vegetation}} 
        & \rotatebox{90}{\textcolor{trunk}{$\blacksquare$} trunk\classfreq{trunk}} 
        & \rotatebox{90}{\textcolor{terrain}{$\blacksquare$} terrain\classfreq{terrain}} 
        & \rotatebox{90}{\textcolor{person}{$\blacksquare$} person\classfreq{person}} 
        & \rotatebox{90}{\textcolor{bicyclist}{$\blacksquare$} bicyclist\classfreq{bicyclist}} 
        & \rotatebox{90}{\textcolor{motorcyclist}{$\blacksquare$} motorcycl.\classfreq{motorcyclist}} 
        & \rotatebox{90}{\textcolor{fence}{$\blacksquare$} fence\classfreq{fence}} 
        & \rotatebox{90}{\textcolor{pole}{$\blacksquare$} pole\classfreq{pole}} 
        & \rotatebox{90}{\textcolor{traffic-sign}{$\blacksquare$} traf.-sign\classfreq{trafficsign}} 
\\
\midrule
            LMSCNet~\citeyearpar{roldaoLMSCNetLightweightMultiscale2020} & L  &  31.4 & 7.1 & 46.7 & 19.5 & 13.5 & 3.1 & 10.3 & 14.3 & 0.3 & 0.0 & 0.0 & 0.0 & 10.8 & 0.0 & 10.4 & 0.0 & 0.0 & 0.0 & 5.4 & 0.0 & 0.0   \\
        3DSketch~\citeyearpar{chen3DSketchawareSemantic2020} & C\&D & 26.9 & 6.2 & 37.7 & 19.8 & 0.0 & 0.0 & 12.1 & 17.1 & 0.0 & 0.0 & 0.0 & 0.0 & 12.1 & 0.0 & 16.1 & 0.0 & 0.0 & 0.0 & 3.4 & 0.0 & 0.0  \\
        JS3C-Net~\citeyearpar{yanSparseSingleSweep2021} & L & 34.0 & 9.0 & 47.3 & 21.7 & 19.9 & 2.8 & 12.7 & 20.1 & 0.8 & 0.0 & 0.0 & 4.1 & 14.2 & 3.1 & 12.4 & 0.0 & 0.2 & 0.2 & 8.7 & 1.9 & 0.3   \\
        MonoScene~\citeyearpar{caoMonoSceneMonocular3D2022} & C & 34.2 & 11.1 & 54.7 & 27.1 & 24.8 & 5.7 & 14.4 & 18.8 & 3.3 & 0.5 & 0.7 &  \underline{4.4} & 14.9 & 2.4 & \underline{19.5} & 1.0 & 1.4 & 0.4 & 11.1 & 3.3 & 2.1  \\

        SurroundOcc~\citeyearpar{weiSurroundOccMultiCamera3D2023} & C & {34.7} & {11.9} & \underline{56.9} & \underline{28.3} & \underline{30.2} & \underline{6.8} & {15.2} & {20.6} & 1.4 & \underline{1.6} & \underline{1.2} & \underline{4.4} & {14.9} & {3.4} & 19.3 & \underline{1.4} & \underline{2.0} & 0.1 & \underline{11.3}& {3.9} & {2.4} \\ 

        SCPnet~\citeyearpar{xiaSCPNetSemanticScene2023} & L & \underline{{56.1}} & \textbf{{36.7}} & \textbf{68.5} & \textbf{49.8} & \textbf{51.3} & \textbf{30.7} & \textbf{38.8} & \textbf{46.4} & \textbf{13.8} & \textbf{33.2} & \textbf{34.9} & \textbf{29.1} & \textbf{46.4} & \textbf{40.1} & \textbf{48.7} & \textbf{28.2} & \textbf{24.7} & {34.9} & \textbf{44.7} & \textbf{40.4} & \textbf{25.1} \\

        L-CONet~\citeyearpar{wangOpenOccupancyLargeScale2023} & L & 55.6 & 4.7 & 20.2 & 6.1 & 1.6 & 0.0 & 1.7 & 2.3 & 0.0 & 0.0 & 0.0 & 0.0 & 0.5 & 0.0 & 6.5 & 0.0 & 0.0 & \underline{49.2} & 1.0 & 0.1 & 0.0 \\ 

\midrule


MergeOcc (D.M.)(ours) & L & 36.7 & 2.0 & 1.6 & 1.1 & 0.0 & 0.0 & 0.1 & 0.3 & 0.0 & 0.0 & 0.0 & 0.0 & 1.7 & 0.0 & 0.9 & 0.0 & 0.0 & 31.5 & 0.2 & 0.0 & 0.0  \\

MergeOcc (Tr.both)(ours) & L & \textbf{{70.1}} & \underline{{14.5}} & {35.4} & {19.3} & {3.4} & {0.0} & \underline{21.8} & \underline{32.9} & \underline{9.5} & {0.0} & {0.5} & {1.0} & \underline{27.3} & \underline{7.9} & {18.2} & {0.5} & {0.3} & \textbf{63.7} & {7.3} & \underline{19.4} & \underline{6.2}  \\
        
\bottomrule
\end{tabular}}
\caption{\textbf{3D Semantic occupancy prediction results on SemanticKITTI.} We select the same metric as OpenOccupancy-nuScenes.}
\label{tab:semkitti result}
\end{table*}

\subsection{Supplementary Experiments Results}
\label{sec:appendixc}
\subsubsection{3D Occupancy Prediction Results on SemanticKITTI}
It is worth mentioning that all the experimental results on both two datasets were obtained by taking the average value of three experiments.
The class-wise 3D occupancy prediction results for the SemanticKITTI dataset are presented in \cref{tab:semkitti result}. A detailed comparison between MergeOcc and previous methods has been provided in the \cref{sec:Exp}.

\subsubsection{Further Ablation about R.A.}
Ablation experiments indicate that the point cloud range alignment (R.A.) module plays an important role in improving the performance of MergeOcc. To elucidate the underlying causes, we conducted additional ablation experiments focusing on R.A., with results detailed in \cref{tab:RA effect}. These findings demonstrate that R.A. is beneficial exclusively within the Multiple Datasets Training (MDT) paradigm. Conversely, under the single dataset training paradigm, R.A. is detrimental due to inherent data limitations. The principal performance improvements are attributed to the \textbf{learning afforded by larger and more diverse datasets}, with R.A. \textbf{serving as the bridge} to address existing gaps. This outcome highlights the superiority and critical importance of the MDT paradigm.

\begin{table}[!htb]
\begin{minipage}[t]{0.495\textwidth} 
		\centering
		\begin{tabular}[b]{c|c|c|c}
			\toprule
			Model & Dataset &  IoU &  mIoU 
			\\
			\midrule
			L-CoNet(baseline) & OO-nu & 30.9 & 15.8 \\
			MergeOcc(R.A.) & OO-nu & 27.4{$(-3.5)$} & 12.5{$(-3.3)$} \\
			MergeOcc(R.A.) & Both & 38.2{$(+7.3)$} & 21.3{$(+5.5)$}
            \\
			\bottomrule
		\end{tabular}
      \caption{{R.A. benefits only under the MDT paradigm}}
	\label{tab:RA effect}
\end{minipage}
\end{table}

\subsection{Scope of Using Geometric Alignment}
\label{sec:appendixd}
We investigate the optimal range for geometric statistical data alignment, referred to as the utilization range of the dataset-specific norm layer. We categorize the settings into two distinct configurations: (1) employing geometric alignment only at the backbone and (2) substituting all norm layers in the network with geometric alignment. The experimental results are delineated in \cref{tab: norm range}, prompting us to advocate for using geometric alignment solely for the backbone in the network.

\begin{table*}[htbp]
\centering
\begin{small}
{
    \begin{tabular}{c|c|c|c|c|c}
        \bottomrule[1pt]
        \multirow{2}{*}{Trained on} & \multirow{2}{*}{Method}  & \multicolumn{2}{c|}{Tested on OpenOccupancy-nuScenes} & \multicolumn{2}{c}{Tested on SemanticKITTI} \\
        \cline{3-6}
        &  & Geometric IoU & Semantic mIoU & Geometric IoU & Semantic mIoU \\
        \hline
        \multirow{2}{*}{SK+OO-nu} & MergeOcc  (G.A. for all)   & 37.0  &  21.2  &  68.7  & 14.2  \\
        & MergeOcc  (G.A. for backbone)  & 38.4{($+$1.4)}  & 21.4{($+$0.2)}   & 69.9{($+$1.2)} & 14.5{($+$0.3)}  \\
        \toprule[1pt]
    \end{tabular}
}
\caption{Experimental results of different scopes of geometric alignment.}
\label{tab: norm range}
\end{small}
\end{table*}

\subsection{Coarse to Fine Stage}
\label{sec:appendixe}
We use the geometric coarse to fine query to upsample the initial output, inspired by OpenOccupancy~\cite{wangOpenOccupancyLargeScale2023}.

Specifically, the coarse occupancy $O^\mathcal{M}\in\mathbb{R}^{\frac{D}{S}\times \frac{H}{S} \times \frac{W}{S} \times c}$ is first generated by the baseline model, where the occupied voxels $V_{\rm{o}}\in\mathbb{R}^{N_{\rm{o}}\times 3}$ ($N_{\rm{o}}$ is the number of occupied voxels, and 3 denotes the $(x,y,z)$ indices in voxel coordinates) are split as high-resolution occupancy queries $Q_{\rm{H}}\in\mathbb{R}^{N_{\rm{o}}8^{\eta-1}\times 3}$:
\begin{equation}
    Q_{\rm{H}} = \mathcal{T}_{\rm{v}\rightarrow \rm{w}}(\mathcal{F}_{\rm{s}}(V_{\rm{o}}, \eta)),
\end{equation}
where $\mathcal{F}_{\rm{s}}$ is the voxel split function (i.e., for $(x_0,y_0,z_0)$ in $V_{\rm{o}}$, the split indices are $\{x_0+\frac{i}{\eta},y_0+\frac{j}{\eta},z_0+\frac{k}{\eta}\}(i,j,k\in(0,\eta-1))$), $\eta$ is the split ratio (typically set as 4), and $\mathcal{T}_{\rm{v}\rightarrow \rm{w}}$ transforms the voxel coordinates to the world coordinates. 
Subsequently, we transform $Q_{\rm{H}}$ to voxel space to sample geometric features $F^{\mathcal{G}}=\mathcal{G}_{\rm{S}}(F^{\mathcal{F}},\mathcal{T}_{\rm{w}\rightarrow\rm{v}}(Q_{\rm{H}}))$ ($\mathcal{G}_{\rm{S}}$ is the \textit{grid sample} function, $\mathcal{T}_{\rm{w}\rightarrow\rm{v}}$ is the transformation from world coordinates to voxel coordinates). FC layers then regularize the sampled features to produce fine-grained occupancy predictions:
\begin{equation} 
    O^{\rm{g}} = \mathcal{G}_f(\mathcal{G}_f(F^{\mathcal{G}})),
\end{equation}
where $F^{\mathcal{G}}$ are FC layers. Finally, $O^{\rm{g}}$ can be reshaped to the volumetric representation $O^{\rm{vol}}\in\mathbb{R}^{\frac{\eta D}{S}\times \frac{\eta H}{S} \times \frac{\eta W}{S} \times c}$:
\begin{footnotesize}
\begin{equation}
  O^{\rm{vol}}(x,y,z) = \left\{\begin{array}{ll}
    O^{\rm{g}}(\mathcal{T}_{\rm{v}\rightarrow\rm{q}}(x,y,z)) & (x,y,z)\in \mathcal{T}_{\rm{w}\rightarrow\rm{v}}(Q_{\rm{H}})\\
    \text{Empty~Label} & (x,y,z)\notin \mathcal{T}_{\rm{w}\rightarrow\rm{v}}(Q_{\rm{H}}),
    \end{array}\right.
\end{equation}
\end{footnotesize}
where $\mathcal{T}_{\rm{v}\rightarrow\rm{q}}$ transforms the voxel coordinates to indices of the high-resolution query $Q_{\rm{H}}$. For LiDAR-based CONet that without multi-view 2D features, we only sample $Q_{\rm{H}}$ from $F^{\mathcal{L}}$.

\subsection{More Illustration about Semantic Label Mapping}
\label{sec:appendixf}
\subsubsection{Semantic Level differences}
Different datasets have obvious semantic disparities. For instance, SemanticKITTI distinguishes static and dynamic objects (\textbf{e.g.}, `car' vs. `moving-car'), which is crucial for scene understanding. Furthermore, it aggregates certain object categories into broader classes (\textbf{e.g.}, `other-vehicle' encompasses buses and rail vehicles) and details categories for road infrastructure (`road', `parking', `sidewalk') and natural elements (`vegetation', `terrain').   Conversely, nuScenes~\cite{caesarNuScenesMultimodalDataset2020} focuses on distinguishing various vehicle types (\textbf{e.g.}, `car', `truck', `bus') and explicitly categorizes urban infrastructure elements such as `traffic-cone'. It also clearly segregates `driveable surface' from `sidewalk', establishing a distinct boundary between zones that are navigable by vehicles and those meant for pedestrian use.

Besides, the distribution of point clouds varies significantly across datasets, owing to their collection from diverse geographical locales via different LiDARs, as depicted in \cref{tab1}. Such heterogeneity engenders pronounced disparities in road topographies and object dimensions, as exemplified by the visualizations presented in \cref{fig: visualization},\cref{fig: oo-nu visualization} and \cref{fig: sk visualization}.




\subsubsection{Computation of Label Space Learning Algorithm}
\label{sec:prune}

The size of our optimization problem scales linearly in the number of potential merges $|\mathbb{T}|$, which can grow exponentially in the number of datasets.
To counteract this exponential growth, we only consider sets of classes $$\mathbb{T}^\prime = \left\{\vec t \in \mathbb{T} \bigg| \frac{c_{\vec t}}{|\vec t|-1} \le \tau\right\}.$$
For an aggressive enough threshold $\tau$, the number of potential merges $|\mathcal{T}^\prime|$ remains manageable.
We greedily grow $\mathcal{T}^\prime$ by first enumerating all feasible two-class merges ($|\vec t|=2$), then three-class merges, and so on.
The detailed algorithm diagram is shown in Algorithm~\ref{alg:labelspace}.
The time complexity of this greedy algorithm is $O(|\mathcal{T}^\prime|\max_i|\hat L^i|)$.

\begin{algorithm}[!th]
    \caption{Learning a unified label space}
	\label{alg:labelspace}
	\SetAlgoLined
	\SetKwInOut{Input}{Input} \SetKwInOut{Output}{Output} 
    \Input{$\{{\bf o}_i, \hat{\bf l}_i\}_{i=1}^N$: semantic occupancy grids ground truth and labels for each of the N training datasets\\
      $\{\{\tilde{\bf o}_i^{(j)}, \tilde{\bf l}_i^{(j)}\}_{j=1}^{N}\}_{i=1}^{N}$: predicted semantic occupancy grids with predicted classes in all datasets for each training dataset\\
      $\lambda, \tau$: hyper-parameters for algorithm\\
     }
	\Output{L: unified label space\\
	  $\mathcal{T}$: the transformation from each individual label space to the unified label space}
	// Compute potential merges and merge cost\\
	$\hat L = \bigcup_i \hat L_i$ // Short-hand used to simplify notation\\
	$\mathbb{T}_1 \gets \{(l) | l \in \hat L\}$  // Set of single labels\\
	Compute $c_{\vec t}$ for all single labels $\vec t \in \mathbb{T}$. // 0 for most metrics\\
	\For{$n = 2 \ldots N$}{
	  $\mathbb{T}_n \gets \{\}$\\
	  \For{$\vec t \in \mathbb{T}_{n-1}$} {
    	\For{$l \in \hat L$} {
    	  \If{$l$ and all labels in $\vec t$ are from different datasets}{
    	    compute $c_{\vec t \cup \{l\}}$.\\
    	    \If{$\frac{c_{\vec t \cup \{l\}}}{n-1} \le \tau$}{
    	        Add $\vec t \cup \{l\}$ to $\mathbb{T}_n$.
    	    }
    	  }
    	}
	  }
	}
	$\mathbb{T} \gets \bigcup_{n=1}^N \mathbb{T}_n $ \\
    // Solve the ILP.\\
    $\vec{x} \gets  \text{ILP\_solver}(c, \mathbb{T}, \lambda)$ // Solve equation (8).\\
    Compute $L, \mathcal{T}$ from $\vec{x}$ \\
    \textbf{Return}: $L, \mathcal{T}$
\end{algorithm}

\subsubsection{Sequential Paradigm to Add New Datasets to the Unified Label Space}
While the aspiration is to maintain large and comprehensive training domains and label spaces, practical scenarios often necessitate the inclusion of more fine-grained labels or specific testing domains. Upon establishing a unified label space from an existing set of training datasets, a straightforward label space expansion algorithm is employed to facilitate the addition of further datasets and labels after the unified model has been trained. 

We adopt a sequential optimization paradigm. Specifically, we execute the unified model on the already merged dataset and train a domain-specific perception model on the new domain. Subsequently, the label space learning algorithm is applied to generate a new unified label space. This approach mitigates the computational overhead associated with merging more than two datasets.

\subsection{Challenges of Merging Datasets}
\label{sec:appendixg}
\subsubsection{Dataset Introduction.}
Following the practice of popular scene completion models~\cite{wangOpenOccupancyLargeScale2023,weiSurroundOccMultiCamera3D2023,xiaSCPNetSemanticScene2023}
, our experiments are conducted on two prominent LiDAR semantic occupancy prediction benchmarks, namely,    OpenOccupancy-nuScenes~\cite{wangOpenOccupancyLargeScale2023} and SemanticKITTI~\cite{behleySemanticKITTIDatasetSemantic2019}.

SemanticKITTI comprises annotated outdoor LiDAR scans with 21 semantic labels, organized into 22 point cloud sequences. Sequences 00 to 10, 08, and 11 to 21 are designated for training, validation, and testing, respectively. From these, 19 classes are selected for training and evaluation after merging classes with distinct motion statuses and removing classes with sparse points.

As for OpenOccupancy-nuScenes, it is a large-scale occupancy prediction dataset deriving from nuScenes. Since the 3D semantic and 3D detection labels are unavailable in the test set, Wang et al.~\cite{wangOpenOccupancyLargeScale2023} did not provide dense occupancy labels of the unseen test set. Consequently, we utilize the training set for model training and the validation set for evaluation purposes.

\subsubsection{Primary Aspiration and Challenges}
In the quest for highly intelligent automated vehicles, scalability and generalizability emerge as pivotal characteristics in perception models. According to the scaling law, data plays a crucial role in augmenting both performance and generalizability.

However, acquiring vehicle travel data poses significant challenges compared to other forms of visual or textual data. Particularly, obtaining 3D LiDAR data entails substantial expenses. Consequently, existing autonomous driving datasets exhibit limited data volumes, hindering the training of models with requisite scalability and generalization. The conventional single dataset training-and-testing paradigm confines the source data within a delimited domain. Fully exploiting all available 3D data holds promise for mitigating resource expenditure and enhancing performance.

Initially, we have made a lot of attempts to train a vanilla 3D perception model using multiple datasets by directly merging existing 3D datasets, such as merging OpenOccupancy-nuScenes~\cite{wangOpenOccupancyLargeScale2023} and SemanticKITTI~\cite{behleySemanticKITTIDatasetSemantic2019}. However, we found that commonly employed 3D perception models failed to perform satisfactorily across both datasets, as shown in ~\cref{fig:drop}. This inadequacy stems from the substantial disparities inherent in 3D point clouds acquired by diverse LiDARs, as shown in ~\cref{tab1}, rendering previous 3D models incapable of effectively addressing the significant data shift.

Furthermore, in the architectural design of the models to accommodate diverse label spaces, the incorporation of multiple heads within the network is deemed imperative. To facilitate gradient backward propagation across all network parameters and sustain the training process, a balanced distributed group sampler has been designed. This sampler amalgamates data from disparate datasets within each batch, thereby ensuring that data for each batch is sequentially drawn from distinct datasets.

\subsection{Extended Visualization of Occupancy Prediction Results}
\label{sec:appendixh}

\begin{figure*}[ht]
\centering
\includegraphics[width=0.9\linewidth]{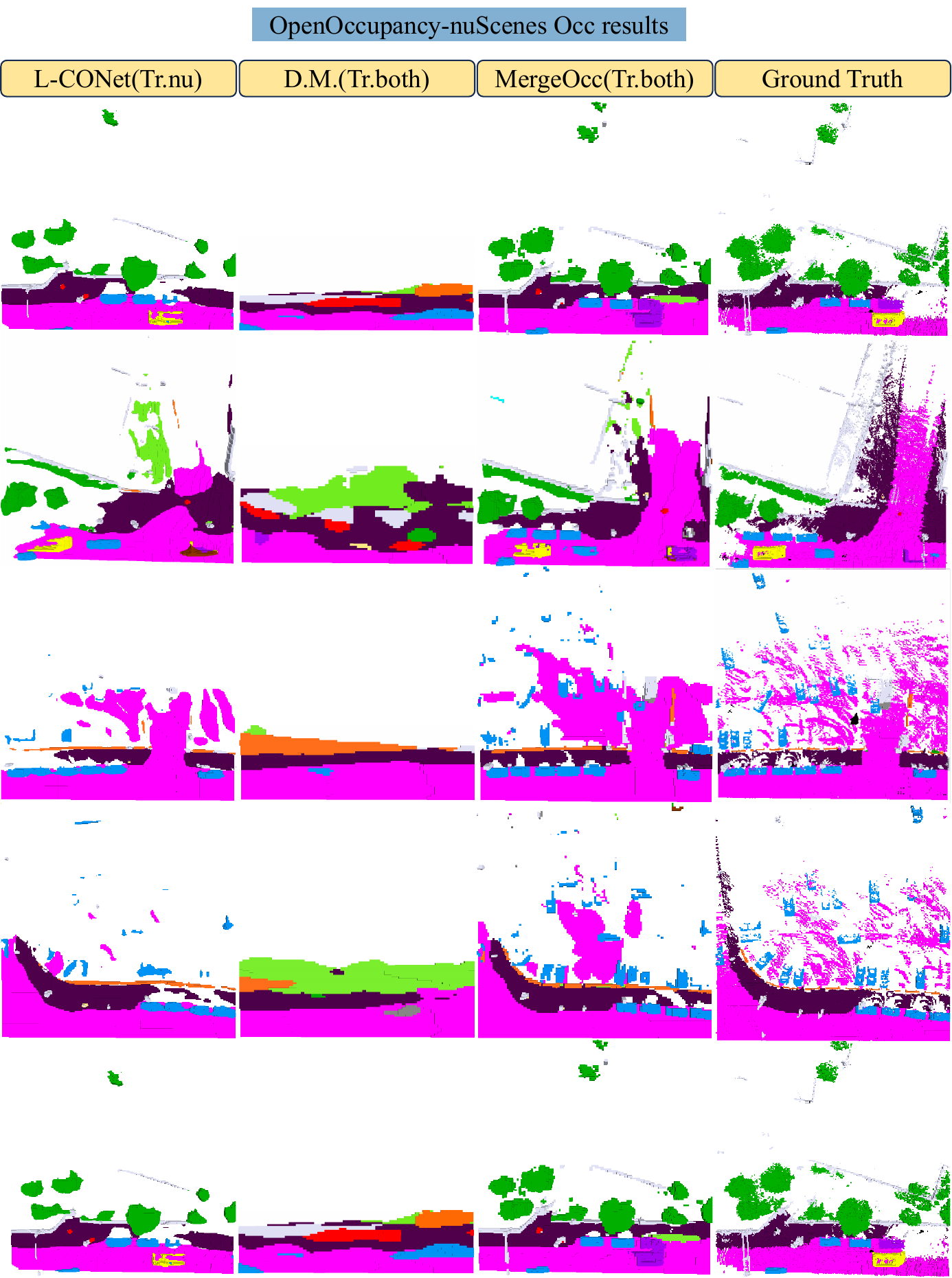}
\caption{Visualizations of occupancy prediction results on OpenOccupancy-nuScenes.}
\label{fig: oo-nu visualization}
\end{figure*}

\begin{figure*}[ht]
\centering
\includegraphics[width=0.9\linewidth]{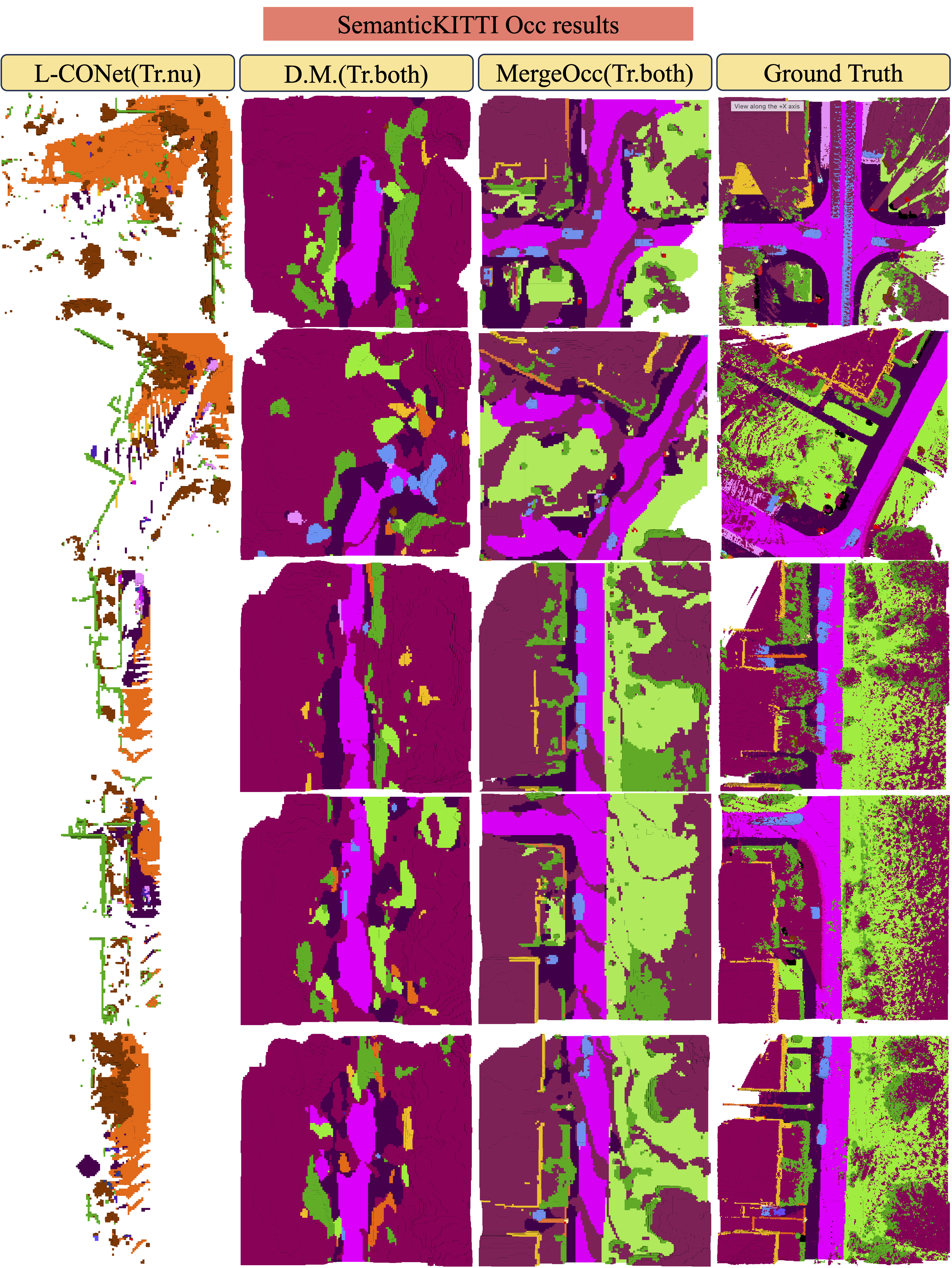}
\caption{Visualizations of occupancy prediction results on SemanticKITTI.}
\label{fig: sk visualization}
\end{figure*}

Further visualization results of the proposed MergeOcc are shown in Figs.~\ref{fig: oo-nu visualization} and~\ref{fig: sk visualization}.

For ~\cref{fig: oo-nu visualization} and~\cref{fig: sk visualization}, we utilize the proposed MergeOcc model trained jointly on OpenOccupancy-nuScenes and SemanticKITTI datasets and showcase the outcomes on the validation sets of these two datasets. These results comprehensively demonstrate our ability to achieve improved occupancy prediction simultaneously for OpenOccupancy-nuScenes and SemanticKITTI datasets using a single perception model.

Furthermore, owing to the denser ground truth annotations provided by the SemanticKITTI dataset, the outcomes yielded by MergeOcc on the OpenOccupancy-nuScenes dataset manifest heightened credibility compared to the ground truth in certain areas, such as the drivable surface and the trunk, thereby underscoring the supremacy of MDT paradigm.

\clearpage
\clearpage


\end{document}